\documentclass[sigconf]{acmart}

\AtBeginDocument{%
  \providecommand\BibTeX{{%
    \normalfont B\kern-0.5em{\scshape i\kern-0.25em b}\kern-0.8em\TeX}}}

\setcopyright{iw3c2w3}
\copyrightyear{2021}
\acmYear{2021}

\acmDOI{10.1145/3442381.3449834}

\acmConference[WWW '21]{Proceedings of the Web Conference 2021}{April 19--23, 2021}{Ljubljana, Slovenia}
\acmBooktitle{Proceedings of the Web Conference 2021 (WWW '21), April 19--23, 2021,
Ljubljana, Slovenia}
\acmPrice{}

\acmISBN{978-1-4503-8312-7/21/04}
\usepackage{amsthm}
\usepackage{tabularx}
\usepackage[ruled,linesnumbered,vlined]{algorithm2e}
\usepackage{caption}
\newtheorem{define}{Definition}[section]

\makeatletter
\newcommand{\algorithmfootnote}[2][\footnotesize]{%
  \let\old@algocf@finish\@algocf@finish
  \def\@algocf@finish{\old@algocf@finish
    \leavevmode\rlap{\begin{minipage}{\linewidth}
    #1#2
    \end{minipage}}%
  }%
}
\makeatother

\begin{document}

\title{Knowledge-Preserving Incremental Social Event Detection via Heterogeneous GNNs}

\author{Yuwei Cao}
\affiliation{
  \institution{University of Illinois at Chicago}
  \streetaddress{851 S. Morgan Street}
  \city{Chicago}
  \state{IL}
  \country{USA}
  \postcode{60607-7053}
}
\email{ycao43@uic.edu}

\author{Hao Peng}
\authornote{This is the corresponding author.}
\affiliation{
  \institution{Beihang University}
  \city{Beijing}
  \country{China}
  \postcode{100083}
}
\email{penghao@act.buaa.edu.cn}

\author{Jia Wu}
\affiliation{
  \institution{Macquarie University}
  \city{Sydney}
  \country{Australia}
}
\email{jia.wu@mq.edu.au}

\author{Yingtong Dou}
\affiliation{
  \institution{University of Illinois at Chicago}
  \streetaddress{851 S. Morgan Street}
  \city{Chicago}
  \state{IL}
  \country{USA}
  \postcode{60607-7053}
}
\email{ydou5@uic.edu}

\author{Jianxin Li}
\affiliation{
  \institution{Beihang University}
  \city{Beijing}
  \country{China}
  \postcode{100083}
}
\email{lijx@act.buaa.edu.cn}

\author{Philip S. Yu}
\affiliation{
  \institution{University of Illinois at Chicago}
  \streetaddress{851 S. Morgan Street}
  \city{Chicago}
  \state{IL}
  \country{USA}
  \postcode{60607-7053}
}
\email{psyu@uic.edu}

\renewcommand{\shortauthors}{Y. CAO et al.}

\begin{abstract}
Social events provide valuable insights into group social behaviors and public concerns and therefore have many applications in fields such as product recommendation and crisis management. The complexity and streaming nature of social messages make it appealing to address social event detection in an incremental learning setting, where acquiring, preserving, and extending knowledge are major concerns. Most existing methods, including those based on incremental clustering and community detection, learn limited amounts of knowledge as they ignore the rich semantics and structural information contained in social data. Moreover, they cannot memorize previously acquired knowledge. In this paper, we propose a novel \underline{K}nowledge-\underline{P}reserving Incremental Heterogeneous \underline{G}raph \underline{N}eural \underline{N}etwork (KPGNN) for incremental social event detection. To acquire more knowledge, KPGNN models complex social messages into unified social graphs to facilitate data utilization and explores the expressive power of GNNs for knowledge extraction. To continuously adapt to the incoming data, KPGNN adopts contrastive loss terms that cope with a changing number of event classes. It also leverages the inductive learning ability of GNNs to efficiently detect events and extends its knowledge from previously unseen data. To deal with large social streams, KPGNN adopts a mini-batch subgraph sampling strategy for scalable training, and periodically removes obsolete data to maintain a dynamic embedding space. KPGNN requires no feature engineering and has few hyperparameters to tune. Extensive experiment results demonstrate the superiority of KPGNN over various baselines. 
\end{abstract}



\maketitle

\section{Introduction}\label{sec:intro}
\begin{figure*}[t]
    \centering
    \includegraphics[width = 17cm]{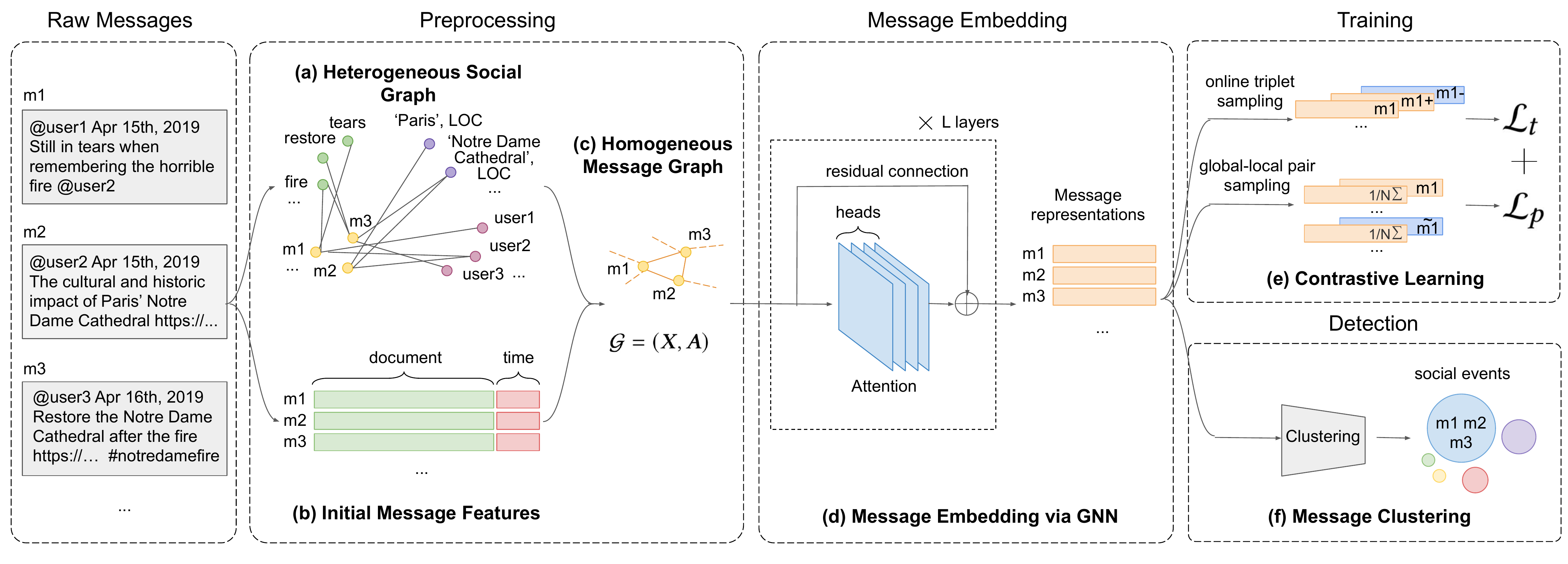}
    \caption{The architecture of the proposed KPGNN model (\textit{best viewed in color}).\textmd{ 
    \textbf{(a)} is a heterogeneous social graph that combines various types of elements contained in the raw messages. Different node colors denote different node types. 
    \textbf{(b)} is the initial feature vectors of the messages. \textbf{(c)} is a homogeneous message graph that incorporates \textbf{(a)} and \textbf{(b)} (detailed in Section \ref{section:preprocessing}). 
    \textbf{(d)} shows a GNN-based encoder that learns representations for the messages in \textbf{(c)}. 
    \textbf{(e)} calculates triplet loss $\mathcal{L}_t$ and global-local pair loss $\mathcal{L}_p$ through contrastive learning. In \textbf{(e)}, two orange bars form a positive pair while one orange bar and one blue bar denote a negative pair. 
    \textbf{(f)} clusters messages into social events.}}
    \label{fig:framework}
    \Description[The architecture of KPGNN]{The architecture contains five components: Raw Messages, Preprocessing, Message Embedding, Training, and Detection. Raw Messages are tweets that include elements such as tweet ids, user ids, timestamps, and text contents. These feed into Preprocessing, which constructs Heterogeneous Social Graph and Initial Message Features, then use them to construct Homogeneous Message Graph, which is fed into Message Embedding. Message Embedding generates message representations, which feed into Training and Detection.}
\end{figure*}

Social events (e.g., Twitter discussions on the Notre-Dame Cathedral fire, as shown in Figure \ref{fig:framework}) highlight significant occurrences in our daily lives, and generally reflect group social behaviors and widespread public concerns. 
Social event detection is very important since it provides valuable insights for us to make timely responses, and therefore has many applications in fields including crisis management, product recommendation, and decision making \cite{zhou2014event,liu2017event}.
In the last decade, social event detection has become the research hot spot in social media mining and has drawn increasing attention from both academia and the industry \cite{peng2019fine,fedoryszak2019real}.

The task of social event detection can be formalized as extracting clusters of co-related messages from social streams (i.e., sequences of social media messages) to represent events (the corresponding methods are categorized as document-pivot, i.e., DP methods \cite{liu2020story,aggarwal2012event,peng2019fine,zhou2014event,hu2017adaptive,zhang2007new}, and are discussed in more detail in Section \ref{section:related_work}). Compared to traditional news and articles, social streams such as Twitter streams are more complex for the following reasons: they are generated in sequential order and are enormous in volume; they contain elements of various types including text, time, hashtags, and implicit social network structures; their contents are short and often contain abbreviations that are not in the dictionary; the semantics of their elements change rapidly. All these characteristics made social event detection a challenging task \cite{fedoryszak2019real, peng2019fine}. 

We argue that the complexity and streaming nature of social messages make it appealing to address the task of social event detection in an incremental learning \cite{rebuffi2017icarl} setting. Incremental learning models are characterized by their abilities of 1) acquiring knowledge from data, 2) preserving previously learned knowledge, and 3) continually adapting to incoming data \cite{rebuffi2017icarl}. 
Existing social event detection methods, however, cannot fully satisfy these requirements of incremental learning.
Traditional methods based on incremental clustering \cite{aggarwal2012event,ozdikis2017incremental,hu2017adaptive,zhang2007new} and community detection \cite{fedoryszak2019real,liu2020story,yu2017ring,liu2020event}, though are capable of detecting events in an online manner, learn limited amounts of knowledge from social data. 
Specifically, they identify events using statistical features such as word frequencies and co-occurrences while ignoring the rich semantics and structural information contained in social streams to some extent. Moreover, these methods have very few parameters in their models. Consequently, they cannot memorize previously learned information, i.e., they forget what they have learned. Motivated by the power of graph neural networks (GNNs) in aggregating structural information and semantics, recent efforts such as \cite{peng2019fine} explore GNN-based social event detection and show promising performance. 
Nevertheless, \cite{peng2019fine} assumes that the entire dataset is available and the output space is fixed. Extending to new data points requires retraining its model from scratch. In a word, the task of incremental social event detection is not yet solved.

In this paper, we address incremental social event detection from a knowledge-preserving perspective, i.e., we design our model to continuously extend its knowledge while detecting events from the incoming social messages.
Nevertheless, such knowledge-preserving incremental social event detection poses significant challenges, which we summarize as follows.
Firstly, as previously mentioned, the model should acquire, preserve, and extend knowledge. 
This requires the model to efficiently organize and process various elements in the social streams for full utilization and effectively interpret these elements to discover underlying knowledge that would help event detection. 
Moreover, the model needs to efficiently update its knowledge accordingly when new messages arrive. 
Given this, continuous training using the new messages is preferred over retraining from scratch.
Secondly, the model needs to handle a changing number of events (classes) that are unknown. Different to the offline scenario where the total number of classes is pre-known and fixed, new events arise continuously in the online scenario. 
It is clear that classification techniques using softmax cross-entropy losses cannot be directly applied. 
Furthermore, predefining the total number of events as a hyperparameter is commonly done \cite{aggarwal2012event,zhou2014event} but undesirable as it introduces additional constraints.
Thirdly, the model needs to scale to large social streams.
As new messages arrive, the model needs to remove the obsolete social messages now and then to maintain a dynamic output space. 
Also, mini-batch training \cite{ruder2016overview} is preferable compared to batch training \cite{ruder2016overview}, as it does not require that the entire training dataset is in memory.

\begin{figure}[t]
    \includegraphics[width = 7.5cm]{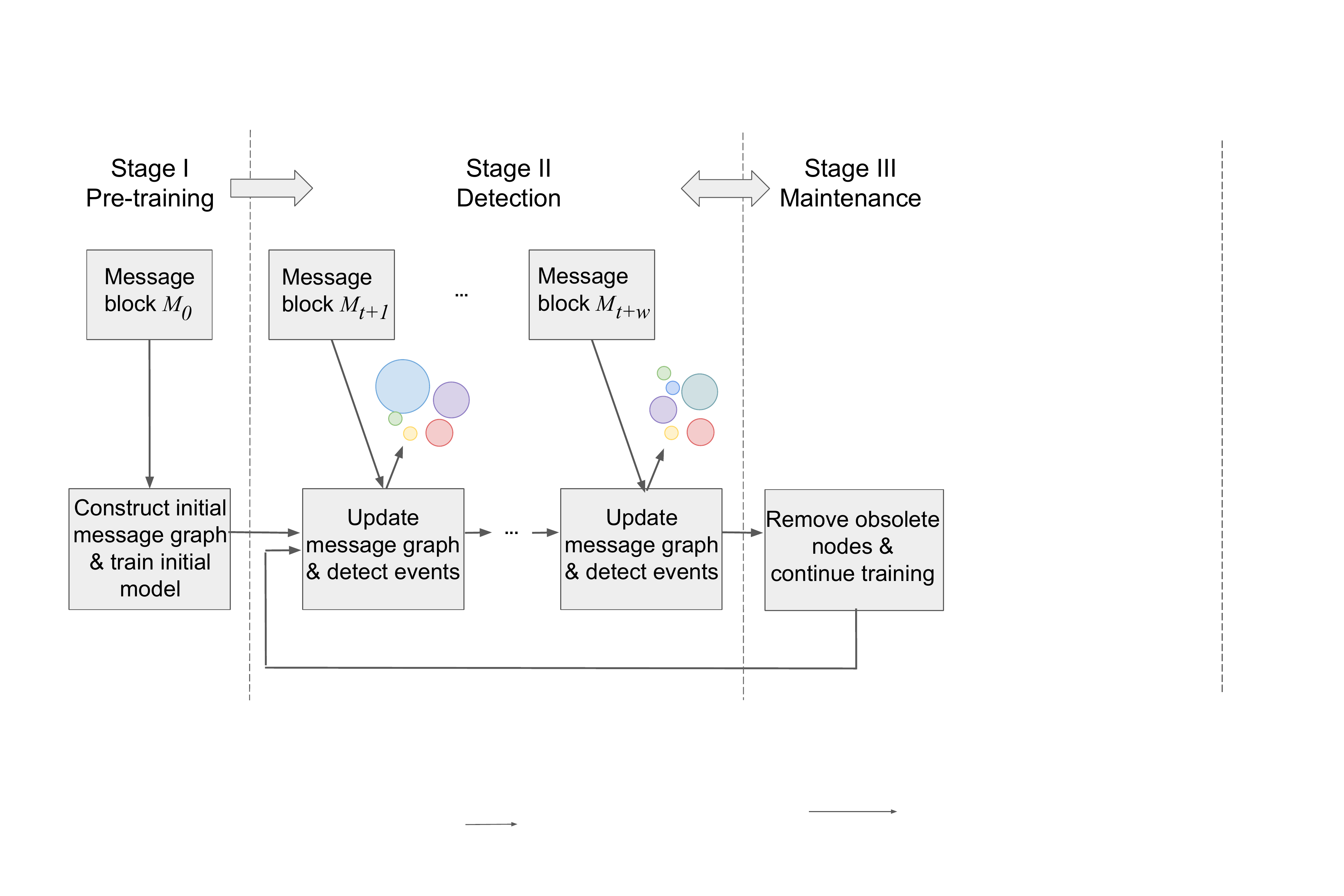}
    \caption{Incremental detection life-cycle of KPGNN. \textmd{\textbf{Stage I} pre-trains an initial KPGNN model. In \textbf{Stage II}, the pre-trained KPGNN model is directly used to detect social events from unseen messages. In \textbf{Stage III}, the KPGNN model is maintained by continuing training using the new messages which arrived in \textbf{Stage II}. The maintained KPGNN model can then be used for the next detection stage. $M_0$, $M_{t+1}$, $M_{t+w}$ denote the input message blocks and $w$ is the window size for maintaining the model. The colored bubbles represent clusters of messages, i.e., social events.}}
    \label{fig:life_cycle}
    \Description[Incremental detection life-cycle of KPGNN]{The life-cycle contains three stages: Pre-training, Detection, and Maintenance. Pre-training takes in a message block to construct an initial message graph and train an initial KPGNN model. Pre-training flows to Detection, which takes in one message block at a time to update the message graph and detect events. The model works for w continuous blocks in the Detection stage before flow to the Maintenance stage, which removes obsolete nodes from the message graph, continues training, and then flows back to the Detection stage.}
\end{figure}

To tackle the above challenges, we propose a novel knowledge-preserving incremental social event detection model based on heterogeneous GNNs. 
We name our model the \underline{K}nowledge-\underline{P}reserving Heterogeneous \underline{G}raph \underline{N}eural \underline{N}etwork (KPGNN).
KPGNN employs a document-pivot technique and classifies social messages based on their correlations. 
1) To address the first challenge, i.e., to acquire, preserve, and extend knowledge, we leverage heterogeneous information networks (HINs) \cite{cao2020multi} to organize social stream elements of various types into unified social graphs. 
We then harness the expressive power of GNNs to acquire knowledge from the semantic and structural information contained in the social graphs. 
The GNN parameters, tuned for social event detection purposes, preserve the model's knowledge about the nature of social data. 
As new messages arrive, the social graphs are subject to change. 
To cope with this, we design a life-cycle of KPGNN (shown in Figure~\ref{fig:life_cycle}) to contain a detection stage that directly detects events from previously unseen messages and a maintenance stage that extends the model's knowledge by resuming the training process using the new data. 
Such inference-maintenance design leverages GNNs' inductive learning ability, which, as pointed out by \cite{galke2019can}, is theoretically discussed \cite{hamilton2017inductive,velivckovic2017graph} yet less explored in real-world applications. 
2) To tackle the second challenge, i.e., dynamic event classes, we introduce contrastive learning techniques into the training process. 
Instead of using cross-entropy loss, we design a triplet loss that contrasts positive message pairs with the corresponding negative ones. 
The triplets are constructed in an online manner, as inspired by computer vision studies \cite{schroff2015facenet,hermans2017defense}, to facilitate incremental learning. 
We also introduce an additional global-local pair loss term to better incorporate the graph structure. 
This term is based on contrasting global-local structural information \cite{hjelm2018learning,velickovic2019deep,belghazi2018mine} and does not require class labels. 
3) To address the third challenge, i.e., scale to large social graphs, we periodically remove obsolete messages from the social graphs to keep an up-to-date embedding space.
We also adopt a mini-batch subgraph sampling algorithm \cite{hamilton2017inductive} for scalable and efficient training.

We conduct extensive experiments on large-scale Twitter corpus \cite{mcminn2013building} and event detection corpus \cite{wang2020MAVEN} which are publicly available. The empirical results show that KPGNN achieves better performance compared to various baselines by effectively preserving event-detection-oriented knowledge. We make our code and preprocessed data publicly available \footnote{\url{https://github.com/RingBDStack/KPGNN}}.

We summarize our main contributions as follows:
\begin{itemize}
\item We formalize the task of social event detection in an incremental learning setting.
\item We design a novel heterogeneous GNN-based knowledge-preserving incremental social event detection model, namely KPGNN. KPGNN continuously detects events from the incoming social streams while possessing the power of interpreting complex social data to accumulate knowledge. To the best of our knowledge, we are the first to use GNNs in incremental social event detection.
\item We empirically demonstrate the effectiveness of the proposed KPGNN model.
\end{itemize}

\section{Notations and Problem Formulation}
We first summarize the main notations used in this paper in Table \ref{table:t1}. 
Then we formalize \textit{Social Stream}, \textit{Social Event}, \textit{Social Event Detection}, and \textit{Incremental Social Event Detection} as follows.

\begin{table}[t]
    \caption{Glossary of Notations.}
    \resizebox{\linewidth}{!}{%
    \begin{tabular}{r|l}  
    \toprule
      \textbf{Notation} & \textbf{Description}\\
      \midrule
      $S;M$ & Social stream; Message block   \\ 
      $m$ & Message or message as a node type\\
      $e;E$ & Event; Set of events\\
      $w$ & The window size for maintaining the model\\
      $o;e;u$ & Word; Named entity; User (node types)  \\
      $\boldsymbol{W}_{mk}^{}$ & The adjacency matrix between node type $m$ and $k$\\
      $\mathcal{G}$& Message graph \\
      $N$ & The total number of messages in $\mathcal{G}$\\
      $\boldsymbol{A}$  & The adjacency matrix of $\mathcal{G}$ \\
      $\boldsymbol{X}$  & The initial feature vectors of the messages in $\mathcal{G}$\\
      $\mathcal{E}(\boldsymbol{X},\boldsymbol{A})$&  GNN that embeds the messages in $\mathcal{G}$\\
      $l;L$ & GNN layer number; Total number of layers\\
      $b;B$ & Mini-batch number; Total number of mini-batches\\
      $\{m_b\}$ & A set of messages in the $b$-th mini-batch\\
      $c_1, ..., c_L$ & The number of neighbors sampled in each layer\\
      $\boldsymbol{h}_{m_i}^{(l)}$ & The representation of $m_i$ at the $l$-th layer\\
      $\boldsymbol{h}_{m_i}$ & The final representation of $m_i$\\
      $m_i+$ & A message in the same class as $m_i$\\
      $m_i-$ & A message that is not in the same class as $m_i$\\
      $\boldsymbol{s}$ & The summary vector of $\mathcal{G}$\\
      $\tilde{ \boldsymbol{h}}_{m_i}$ & The corrupted representation of $m_i$\\
      $\mathcal{L}_t$ & Triplet loss\\
      $\mathcal{L}_p$ & Global-local pair loss\\
      \bottomrule
    \end{tabular}}\label{table:t1}
\end{table}

\begin{define}
A \textbf{social stream} $S = M_0, ..., M_{i-1}, M_i, ...$ is a continuous and temporal sequence of blocks of social messages, where $M_i$ is a {\itshape message block} that contains all the messages which arrive during time period $[t_{i}, t_{i+1})$. 
We denote a {\itshape message block} $M_i$ as $M_i = \{m_j|1\leq j \leq{|M_i|}\}$, where $|M_i|$ is the total number of messages contained in $M_i$, and $m_j$ is one message. 
We denote a {\itshape social message} $m_j$ as $m_j = \{d_j, u_j, t_j\}$, where $d_j$, $u_j$, and $t_j$ stand for the associated text document, users (sender and mentioned users), and timestamp of $m_j$, respectively.
\end{define}

\begin{define}
\label{def:social_event}
A \textbf{social event} $e = \{m_i|1\leq i \leq{\left\lvert e \right\lvert}\}$ is a set of correlated social messages that discuss the same real-world happening. 
Note that we assume each social message belongs to at most one event. 
\end{define}

\begin{define}
Given a message block $M_i$, a \textbf{social event detection} algorithm learns a model $f(M_i;\theta) = E_i$, such that $E_i = \{e_k|1\leq k \leq{\left\lvert E_i \right\lvert}\}$ is a set of events contained in $M_i$. 
Here, $\theta$ denotes the parameter of $f$.
\end{define}

\begin{define}
\label{def:incremental_social_event_detection}
Given a social stream $S$, an \textbf{incremental social event detection} algorithm learns a sequence of event detection models $f_0, ..., f_{t-w}, f_t, ...$, such that $f_t(M_i;\theta_t,\theta_{t-w}) = E_i$ for all message blocks in $\{M_i|t+1\leq i \leq t+w\}$. 
Here, $E_i = \{e_k|1\leq k \leq{\left\lvert E_i \right\lvert}\}$ is a set of events contained in message block $M_i$, $w$ is the window size for updating the model, $\theta_t$ and $\theta_{t-w}$ are the parameters of $f_t$ and $f_{t-w}$, respectively. 
Note that $f_t$ extends the knowledge of its predecessor $f_{t-w}$ by depending on $\theta_{t-w}$. 
In particular, we call $f_0$ which extends no previous model as the {\itshape initial model}.
\end{define}

\section{Methodology}\label{sec:method}
This section introduces our proposed KPGNN model. Section \ref{section:incremental} introduces the life-cycle of KPGNN to give the big picture of how KPGNN operates incrementally. 
Sections \ref{section:preprocessing}-\ref{section:prediction} zoom into the components of KPGNN, which are designed with the aims of incremental knowledge acquiring and preserving. Section \ref{section:complexity} analyzes the time complexity of KPGNN.

\begin{algorithm}[t]
\SetAlgoVlined
\KwIn{A social stream $S = M_0, M_1, M_2, ...$, available labels*, window size $w$, the number of layers $L$, and the number of mini-batches $B$.}
\KwOut{Sets of events: $E_0, E_1, E_2, ...$.}
\For{$t = 0, 1, 2, ...$} {
    \eIf{$t==0$}{
       $\mathcal{G} \gets$ construct initial message graph (Section \ref{section:preprocessing})
    }{
       $\mathcal{G} \gets$ update $M_t$ into message graph (Section \ref{section:preprocessing})
    }
    \If(\tcp*[h]{Detect events from $M_t$}){$t!=0$}{
       \For{$l = 1, 2, ..., L$} {
            $\boldsymbol{h}_{m_i}^{(l)} \gets$ Eq. (\ref{equation:propogation}), $ \forall m_i \in M_t$
       }
       $\boldsymbol{h}_{m_i} \gets \boldsymbol{h}_{m_i}^{(L)}, \forall m_i \in M_t$
       
       $E_t \gets$ message clustering (Section \ref{section:prediction})
    }
    \If(\tcp*[h]{Pre-train or maintain model}){$t\%w==0$}{
        \If{$t!=0$}{
            $\mathcal{G} \gets$ remove obsolete messages (Section \ref{section:preprocessing})
        }
        \For(\tcp*[h]{Train in mini-batches}){$b = 1, 2, ..., B$} {
            $\{m_b\} \gets$ sample a mini-batch of messages from $\mathcal{G}$ (Section \ref{section:training})
            
            \For{$l = 1, 2, ..., L$} {
                $\boldsymbol{h}_{m_i}^{(l)} \gets$ Eq. (\ref{equation:propogation}), $ \forall m_i \in \{m_b\}$
            }
            $\boldsymbol{h}_{m_i} \gets \boldsymbol{h}_{m_i}^{(L)}, \forall m_i \in \{m_b\}$
            
            $T \gets$ triplet sampling $\forall m_i \in \{m_b\}$(Section \ref{section:training})
            
            $\mathcal{L}_t \gets$ Eq. (\ref{equation:triplet_loss}), $\forall m_i \in \{m_b\}$
            
            $\boldsymbol{s},\tilde{ \boldsymbol{h}}_{m_i} \gets$ calculate summary and corrupted representations $\forall m_i \in \{m_b\}$(Section \ref{section:training})
            
            $\mathcal{L}_p \gets$ Eq. (\ref{equation:pair_loss}), $\forall m_i \in \{m_b\}$
            
            Back-propagation to update parameters;
        }
    }
}

 \caption{\textbf{KPGNN:} Knowledge-Preserving Heterogeneous Graph Neural Network}
 \algorithmfootnote{*Labels are used for pre-training and maintenance; full labeling is not required (see Section \ref{section:training} for details).}
 \label{algorithm:KPGNN}
\end{algorithm}

\subsection{Continuous Detection Framework}
\label{section:incremental}
KPGNN follows Definition \ref{def:incremental_social_event_detection} and operates incrementally. 
Figure \ref{fig:life_cycle} and Algorithm \ref{algorithm:KPGNN} depict the working process of KPGNN. 
As shown in Figure \ref{fig:life_cycle}, the life-cycle of KPGNN comprises three stages, i.e., pre-training, detection, and maintenance. 
In the pre-training stage, we construct an initial message graph (detailed in Section \ref{section:preprocessing}) and train an initial model (Sections \ref{section:embedding} and \ref{section:training}). 
In the detection stage, we update the message graph with the input message block (Section \ref{section:preprocessing}) and detect events (Section \ref{section:prediction}).
The current KPGNN model works on a continuous series of blocks before entering the maintenance stage. 
In the maintenance stage, we remove obsolete messages from the message graph (Section \ref{section:preprocessing}) and resume model training using data that arrived in the last window.
The maintenance stage allows the model to forget obsolete knowledge (we experiment on different forgetting strategies as detailed in Section \ref{section:maintenance_eval}) and equips the model with the latest knowledge. 
The maintained model can then be used for detection in the next window. 
In this manner, KPGNN continuously adapts to the incoming data to detect new events and update the model's knowledge.

\subsection{Heterogeneous Social Message Modeling}
\label{section:preprocessing}
During preprocessing, we aim to 1) fully leverage the social data by extracting different types of informative elements from the messages, and 2) organize the extracted elements in a unified manner to facilitate further processing. 
We leverage heterogeneous information networks (HINs) \cite{peng2019fine,cao2020multi} for these purposes. 
A HIN is a graph that contains more than one type of nodes and edges. 
Figure~\ref{fig:framework}~(a) shows an example of HIN. 
Given a message $m_i$, we extract a set of named entities \footnote{\url{https://spacy.io/api/annotation##section-named-entities}} and words (with very common and very rare words filtered out) from its document. 
The extracted elements, together with a set of users associated with $m_i$ and $m_i$ itself, are added as nodes into a HIN. 
We add edges between $m_i$ and its elements. 
For example, in Figure \ref{fig:framework} (a), from $m_1$, we can extract tweet node ``$m_1$'', word nodes including ``fire'' and ``tears'' (for simplicity, the figure only shows two words, while there are more to be extracted), and user nodes including ``user1'' and ``user2''. 
We add edges between ``$m_1$'' and the other nodes. 
We repeat the same process for all the messages, with repetitive nodes merged. 
Eventually, we get a heterogeneous social graph containing all the messages and their elements of different types. 
We denote the node types, i.e., message, word, named entity, and user as $m$, $o$, $e$, and $u$, respectively.

Existing heterogeneous GNNs \cite{hu2020heterogeneous,yun2019graph,wang2019heterogeneous,zhang2019heterogeneous,yang2020heterogeneous} typically retain heterogeneous node types throughout their models to learn the representations for all the nodes. However, KPGNN, as a document-pivot model, focuses on learning the correlations between messages and therefore we adopt a different design and map the heterogeneous social graph into a homogeneous message graph as shown in Figure \ref{fig:framework} (c). The homogeneous message graph only contains message nodes and there are edges between messages that share some common elements. 
Through mapping, the homogeneous message graph preserves the message correlations encoded by the heterogeneous social graph. Specifically, the mapping process follows:
\begin{equation}
\boldsymbol{A}_{i,j} = min\Big\{\big[\sum_{k}\boldsymbol{W}_{mk}^{}\cdot \boldsymbol{W}_{mk}^\intercal\big]_{i,j}, 1\Big\}, k\in\{o,e,u\}.
\end{equation}
Here, $\boldsymbol{A}\in\{0,1\}^{N\times N}$ is the adjacency matrix of the homogeneous message graph, where $N$ is the total number of messages in the graph. 
$\cdot_{i,j}^{}$ stands for the matrix element at the $i$-th row and the $j$-th column, $k$ denotes a node type. 
$\boldsymbol{W}_{mk}^{}$ is a submatrix of the adjacency matrix of the heterogeneous social graph that contains rows of type $m$ and columns of type $k$. 
$\cdot_{}^{\intercal}$ stands for matrix transpose, and $min\big\{,\big\}$ takes the smaller between its two operands.
If messages $m_i$ and $m_j$ link to some common type $k$ nodes, $\big[\boldsymbol{W}_{mk}^{}\cdot \boldsymbol{W}_{mk}^\intercal\big]_{i,j}$ will be greater than or equal to one, and $\boldsymbol{A}_{i,j}$ will be equal to one.

To leverage the semantics and temporal information in the data, we construct feature vectors of the messages, as shown in Figure \ref{fig:framework} (b). 
Specifically, document features are calculated as an average of the pre-trained word embeddings \cite{mikolov2013efficient} of all the words in the documents. 
Temporal features are calculated by encoding the timestamps: we convert each timestamp to OLE date, whose fractional and integral components form a 2-d vector. 
We then perform a message-wise concatenation of the two. 
The resulting initial feature vectors, denoted as $\boldsymbol{X} = \{\boldsymbol{x}_{m_i}\in \mathbb{R}^{d}|1\leq i \leq N\}\}$, where $\boldsymbol{x}_{m_i}$ is the initial feature vector of $m_i$ and $d$ is the dimension, are associated with the corresponding message nodes. 
We denote the homogeneous message graph as $\mathcal{G} = (\boldsymbol{X},\boldsymbol{A})$.

Note that $\mathcal{G}$ is not static. 
When a new message block arrives for detection (shown in Figure \ref{fig:life_cycle} stage II), we update $\mathcal{G}$ by inserting the new message nodes, their linkages with the existing message nodes, and the linkages within themselves into $\mathcal{G}$.
Similarly, we periodically remove obsolete message nodes and edges associated with them from $\mathcal{G}$ (shown in Figure \ref{fig:life_cycle} stage III). We empirically compare different update-maintenance strategies in Section \ref{section:maintenance_eval}.

\subsection{Knowledge-Preserving Incremental Message Embedding}
\label{section:embedding}
To study the correlations between messages in a knowledge-preserving manner, we leverage GNNs to learn message representations. As shown in Figure \ref{fig:framework} (d), we train a GNN encoder $\mathcal{E}:\mathbb{R}^{N\times d}\times \{0,1\}^{N\times N}\rightarrow \mathbb{R}^{N\times d'}$, such that $\mathcal{E}(\boldsymbol{X},\boldsymbol{A}) = \{\boldsymbol{h}_{m_i}\in \mathbb{R}^{d'}|1\leq i \leq N\}$, where $\boldsymbol{h}_{m_i}$ represents the high-level representation of message $m_i$. $\mathcal{E}$ contains $L$ layers and the layer-wise propagation follows:

\begin{equation}
\boldsymbol{h}_{m_i}^{(l)}\leftarrow \overset{heads}{\mathbin\Vert}\Big(\boldsymbol{h}_{m_i}^{(l-1)} \oplus \underset{\forall {m_j}\in \mathcal{N}(m_i)}{Aggregator}\big(Extractor(\boldsymbol{h}_{m_j}^{(l-1)})\big)\Big).
\label{equation:propogation}
\end{equation}
Here, $\boldsymbol{h}_{m_i}^{(l)}$ is the representation of $m_i$ at the $(l)$-th GNN layer, and $\boldsymbol{h}_{m_i}^{(0)} = \boldsymbol{x}_{m_i}$. 
$\mathcal{N}(m_i)$ denotes a set of neighbors of $m_i$ according to $\boldsymbol{A}$. 
$\oplus$ stands for an aggregation, e.g., summation, of the information contained in its two operands. $\overset{heads}{\mathbin\Vert}$ represents head-wise concatenation \cite{vaswani2017attention}. 
$Extractor(\cdot)$ and $Aggregator(\cdot)$ \cite{hamilton2017inductive} are designed differently in different GNNs. 
The former extracts useful information from the neighboring messages' representations while the latter summarizes the neighborhood information. We use $\boldsymbol{h}_{m_i}^{(L)}$ as the final representation of $m_i$, i.e., $\boldsymbol{h}_{m_i}$.

In order for KPGNN to work incrementally and embed previously unseen messages, we adopt the graph attention mechanism \cite{velivckovic2017graph} for neighborhood information extraction and aggregation. 
Our $Extractor(\cdot)$ and $Aggregator(\cdot)$ do not assume a fixed graph structure as in \cite{kipf2016semi,grover2016node2vec,xu2017embedding}, instead, they consider the similarities between the representations of the source message and its neighboring messages. 
In this way, KPGNN handles evolving message graphs where new message nodes continuously join in and the model generalizes to even completely unseen message graphs.

KPGNN preserves knowledge: 
the learned representations encode the model's knowledge about the messages, which is a fusion of natural language semantics, temporal information, and the structural information of the homogeneous message graph; 
the learned parameters preserve the model's cognition about the nature of social data and are especially tuned, via contrastive training (discussed in detail in Section \ref{section:training}), for the social event detection purpose.

\subsection{Scalable Training via Contrastive Learning}
\label{section:training}
As new messages continuously arrive, there can be new events that are previously unseen by the model.
Cross-entropy loss functions, though widely adopted by various GNNs \cite{kipf2016semi,velivckovic2017graph}, are no longer applicable. 
We instead construct a contrastive triplet loss that enables KPGNN to differentiate the events without constraining their total number. 
As shown in Figure \ref{fig:framework} (e), for each message $m_i$ (referred to as an anchor message), we sample a positive message $m_i+$ (i.e., a message from the same class) and a negative message $m_i-$ (i.e., a message from a different class) to form a triplet $(m_i,m_i+,m_i-)$. 
The triplet loss function pushes positive messages close to and negative messages far away from anchors and is formalized as:

\begin{equation}
\mathcal{L}_t = \underset{(m_i,m_i+,m_i-)\in T}{\sum}max\big\{\mathcal{D}(\boldsymbol{h}_{m_i},\boldsymbol{h}_{m_i+})-\mathcal{D}(\boldsymbol{h}_{m_i},\boldsymbol{h}_{m_i-})+a,0\big\}.
\label{equation:triplet_loss}
\end{equation}
Here, $\mathcal{D}(,)$ computes the Euclidean distance between two vectors. 
$a \in \mathbb{R}$ is a hyperparameter controlling how farther away the negative messages should be compared to the positive ones. 
$max\big\{,\big\}$ takes the larger between its two operands.
$T$ is a set of triplets sampled in an online manner \cite{hermans2017defense} and we focus on the hard triplets \cite{hermans2017defense}, i.e., triplets that satisfy $\mathcal{D}(\boldsymbol{h}_{m_i},\boldsymbol{h}_{m_i-})<\mathcal{D}(\boldsymbol{h}_{m_i},\boldsymbol{h}_{m_i+})$, as the usage of hard triplets results in faster convergence \cite{hermans2017defense}.

To better incorporate the structural information of the message graph, we construct an additional global-local pair loss that enables KPGNN to discover and preserve the features of similar local structures, as shown in Figure \ref{fig:framework} (e).
Inspired by \cite{velickovic2019deep}, the global-local pair loss takes a noise-contrastive form. 
It maximizes the mutual information between the local message representations and the global summary of the message graph by minimizing their cross-entropy:
\begin{equation}
\mathcal{L}_p = {\frac{1}{N}}\sum_{i=1}^{N}\Big(\log\mathcal{S}(\boldsymbol{h}_{m_i},\boldsymbol{s})+\log\big(1-\mathcal{S}(\tilde{ \boldsymbol{h}}_{m_i},\boldsymbol{s})\big)\Big).
\label{equation:pair_loss}
\end{equation}
Here, $\boldsymbol{s}\in \mathbb{R}^{d'}$ is a summary of the message graph (we simply use the average of all message representations). 
$\tilde{ \boldsymbol{h}}_{m_i}$ is a corrupted representation of $m_i$ learned by $\mathcal{E}(\boldsymbol{\tilde{X}},\boldsymbol{A})$, where $\boldsymbol{\tilde{X}}$ is constructed by the row-wise shuffle of $\boldsymbol{X}$. 
$\mathcal{S}(,)$ is a bilinear scoring function that outputs the probability of its two operands coming from a joint distribution (i.e., being learned from the same graph). 
Note $\mathcal{L}_p$ is applicable to dynamic message graphs and we show in Sections \ref{section:offline_eval} and \ref{section:online_eval} with experiments how $\mathcal{L}_p$ helps improve the performance.
The overall loss of KPGNN is simply the summation of $\mathcal{L}_t$ and $\mathcal{L}_p$.

To make KPGNN scalable to large message graphs, we adopt a mini-batch subgraph sampling \cite{hamilton2017inductive} during training. 
The triplets used in $\mathcal{L}_t$ are constructed from each mini-batch. 
$\boldsymbol{h}_{m_i}$, $\tilde{ \boldsymbol{h}}_{m_i}$ and $\boldsymbol{s}$ in $\mathcal{L}_p$ are also calculated from each subgraph.

It is important to note: 
1) KPGNN, as an incremental model, is not trained once and for all. Instead, we periodically resume the training to keep the model's knowledge up-to-date, as shown in Figure \ref{fig:life_cycle} stage III.
In the maintenance stage, the training does not start from scratch, rather, it is continued based on the previous knowledge (i.e., the existing model parameters) using the new data which arrived during the last time window. 
2) Although the calculation of $\mathcal{L}_t$ needs labels, KPGNN does not require full labeling, as $T$ can be sampled from the labeled messages. 
The unlabeled messages also contribute, as their features and structural information could be aggregated into the representations of the labeled ones through propagation (detailed in Section \ref{section:embedding}). 
The calculation of $\mathcal{L}_p$, on the other hand, does not require any labels. 
Such design suits the real-world scenarios where hashtags can be used as labels and the social streams can be considered as partially labeled.

\subsection{Message Clustering}
\label{section:prediction}
At the detection stage, we cluster the messages based on the learned message representations. Distance-based clustering algorithms such as K-Means and density-based ones such as DBSCAN \cite{ester1996density} can be readily used for clustering the representations.
Of these, \cite{ester1996density} does not require the total number of classes to be specified and therefore suits the need for incremental detection.
This process is shown in Figure \ref{fig:framework} (f). 
KPGNN outputs the resulting message clusters as social events, following Definition \ref{def:social_event}.

\subsection{Time Complexity of KPGNN}
\label{section:complexity}
The overall running time of KPGNN is $O(N_e)$, where $N_e$ is the total number of edges in the message graph. 
Specifically, the running time of constructing the initial message graph (Algorithm \ref{algorithm:KPGNN} line 3) or updating the message graph (Algorithm \ref{algorithm:KPGNN} line 5) is $O(N+N_e)=O(N_e)$, where $N$ is the total number of messages in the message graph. 
The propagation of the GNN encoder $\mathcal{E}$ (Algorithm \ref{algorithm:KPGNN} lines 7-9 and 16-18) takes $O(Ndd'+N_ed')=O(N_e)$, where $d$ and $d'$ are the input and output dimensions of $\mathcal{E}$. 
The mini-batch subgraph sampling (Algorithm \ref{algorithm:KPGNN} line 15) takes $O(\prod_{l=1}^Lc_l)$, where $c_1,...,c_l,...c_L$ are $L$ user-specified constants that define the number of neighbors sampled from the neighborhood of one message in each layer. 
In practice, $\prod_{l=1}^Lc_l\ll N_e$. 
Triplet sampling (Algorithm \ref{algorithm:KPGNN} line 19) takes $O(\sum_{b=1}^{B}{|\{m_b\}|}^2)$, where $|\{m_b\}|$ is the number of messages in the $b$-th batch. 
The corruption of the message graph (Algorithm \ref{algorithm:KPGNN} line 21) takes $O(N)$. It is clear that maintaining a light-weighted message graph helps reduce time consumption and we compare different maintenance strategies in Section \ref{section:maintenance_eval}.

\begin{table*}[t]
\caption{Offline evaluation results on the Twitter dataset. \textmd{The best results are marked in bold and second-best in italic.}}
  \centering
  \begin{tabular}{c|ccccccc|cc}
   \toprule
    Metrics & Word2vec \cite{mikolov2013efficient} & LDA \cite{blei2003latent} & WMD \cite{kusner2015word} & BERT \cite{devlin2018bert} & BiLSTM \cite{graves2005framewise} & PP-GCN \cite{peng2019fine} & EventX \cite{liu2020story} & KPGNN$_t$  &KPGNN \\
    \midrule
    NMI &.44$\pm$.00&.29$\pm$.00&.65$\pm$.00&.64$\pm$.00&.63$\pm$.00&.68$\pm$.02&\textbf{.72$\pm$.00}&.69$\pm$.01&\textit{.70$\pm$.01} \\ 
    AMI  &.13$\pm$.00&.04$\pm$.00&.50$\pm$.00&.44$\pm$.00&.41$\pm$.00 &.50$\pm$.02 &.19$\pm$.00&\textit{.51$\pm$.00} &\textbf{.52$\pm$.01} \\ 
    ARI &.02$\pm$.00&.01$\pm$.00 &.06$\pm$.00&.07$\pm$.00&.17$\pm$.00&.20$\pm$.01 &.05$\pm$.00&\textit{.21$\pm$.01} &\textbf{.22$\pm$.01} \\ 
    \bottomrule
  \end{tabular}
  \label{table:offline_Twitter}
\end{table*}

\begin{table*}[t]
\caption{Offline evaluation results on the MAVEN dataset. \textmd{The best results are marked in bold and second-best in italic.}}
  \centering
  \begin{tabular}{c|ccccccc|cc}
   \toprule
    Metrics & Word2vec \cite{mikolov2013efficient} & LDA \cite{blei2003latent} & WMD \cite{kusner2015word} & BERT \cite{devlin2018bert} & BiLSTM \cite{graves2005framewise} & PP-GCN \cite{peng2019fine} & EventX \cite{liu2020story} & KPGNN$_t$  &KPGNN \\
    \midrule
    NMI &.42$\pm$.00&.35$\pm$.00&.46$\pm$.00&.45$\pm$.00&.44$\pm$.00&.49$\pm$.01&\textbf{.69$\pm$.00}&.51$\pm$.01&\textit{.52$\pm$.01} \\ 
    AMI  &.08$\pm$.00&.04$\pm$.00&.11$\pm$.00&.08$\pm$.00&.06$\pm$.00 &.15$\pm$.01 &.01$\pm$.00&\textit{.19$\pm$.01} &\textbf{.19$\pm$.00} \\ 
    ARI &.02$\pm$.00&.01$\pm$.00 &.04$\pm$.00&.02$\pm$.00&.02$\pm$.00&.06$\pm$.00 &.00$\pm$.00&\textbf{.10$\pm$.00} &\textbf{.10$\pm$.00} \\ 
    \bottomrule
  \end{tabular}
  \label{table:offline_MAVEN}
\end{table*}

\begin{table*}[t]
\caption{The statistics of the social stream.}
  \centering
    \begin{tabular}{c|ccccccccccc}
   \toprule
    Blocks & $M_0$ & $M_1$ & $M_2$ & $M_3$ & $M_4$ & $M_5$ & $M_6$ & $M_7$ & $M_8$ & $M_9$ & $M_{10}$\\
    \midrule
    $\#$ of messages & $20,254$ & $8,722$ & $1,491$ & $1,835$ & $2,010$ & $1,834$ & $1,276$ & $5,278$ & $1,560$ & $1,363$ & $1,096$ \\ 
    \toprule
    Blocks & $M_{11}$ & $M_{12}$ & $M_{13}$ & $M_{14}$ & $M_{15}$ & $M_{16}$ & $M_{17}$ & $M_{18}$ & $M_{19}$ & $M_{20}$ & $M_{21}$ \\ 
    \midrule
    $\#$ of messages & $1,232$ & $3,237$ & $1,972$ & $2,956$ & $2,549$ & $910$ & $2,676$ & $1,887$ & $1,399$ & $893$ & $2,410$ \\ 
    \bottomrule
  \end{tabular}
  \label{table:social_stream}
\end{table*}

\begin{table*}[t]
	\addtolength{\tabcolsep}{+8pt}
\caption{Incremental evaluation NMIs. \textmd{The best results are marked in bold and second-best in italic.}}
  \centering
    \begin{tabular}{c|ccccccc}
      \toprule
    Blocks & $M_1$ & $M_2$ & $M_3$ & $M_4$ & $M_5$ & $M_6$ & $M_7$ \\
    \midrule
    Word2vec \cite{mikolov2013efficient} &.19$\pm$.00 & .50$\pm$.00 & .39$\pm$.00 & .34$\pm$.00 & .41$\pm$.00 & .53$\pm$.00 & .25$\pm$.00 \\
    LDA \cite{blei2003latent} & .11$\pm$.00 & .27$\pm$.01 & .28$\pm$.00 & .25$\pm$.00 & .26$\pm$.00 & .32$\pm$.00 & .18$\pm$.01 \\
    WMD \cite{kusner2015word}& .32$\pm$.00 & .71$\pm$.00 & .67$\pm$.00 & .50$\pm$.00 & .61$\pm$.00 & .61$\pm$.00 & .46$\pm$.00  \\
    BERT \cite{devlin2018bert}&
    .36$\pm$.00 & .78$\pm$.00 & .75$\pm$.00 & .60$\pm$.00 & .72$\pm$.00 & .78$\pm$.00 & .54$\pm$.00 \\
    BiLSTM \cite{graves2005framewise} &
    .24$\pm$.00 & .50$\pm$.00 & .39$\pm$.00 & .40$\pm$.00 & .41$\pm$.00 & .50$\pm$.00 & .33$\pm$.00 \\
    PP-GCN \cite{peng2019fine} &.23$\pm$.00 & .57$\pm$.02 & .55$\pm$.01 & .46$\pm$.01 & .48$\pm$.01 & .57$\pm$.01 & .37$\pm$.00 \\
    EventX \cite{liu2020story}& .36$\pm$.00 & .68$\pm$.00 & .63$\pm$.00 & .63$\pm$.00 & .59$\pm$.00 & .70$\pm$.00 & .51$\pm$.00 \\
    \hline
    KPGNN$_t$&\textit{.38$\pm$.01} & \textit{.78$\pm$.01} & \textbf{.77$\pm$.00} & \textbf{.68$\pm$.01} & \textbf{.73$\pm$.01} & \textit{.81$\pm$.00} & \textit{.54$\pm$.01} \\
    KPGNN &\textbf{.39$\pm$.00} & \textbf{.79$\pm$.01} & \textit{.76$\pm$.00} & \textit{.67$\pm$.00} & \textbf{.73$\pm$.01} & \textbf{.82$\pm$.01} & \textbf{.55$\pm$.01} \\ \midrule
    \toprule
    Blocks & $M_8$ & $M_9$ & $M_{10}$ & $M_{11}$& $M_{12}$ & $M_{13}$ & $M_{14}$ \\
    \midrule
    Word2vec \cite{mikolov2013efficient} &.46$\pm$.00 & .35$\pm$.00 & .51$\pm$.00 & .37$\pm$.00 & .30$\pm$.00 & .37$\pm$.00 & .36$\pm$.00 \\
    LDA \cite{blei2003latent} & .37$\pm$.01 & .34$\pm$.00 & .44$\pm$.01 & .33$\pm$.01 & .22$\pm$.01 & .27$\pm$.00 & .21$\pm$.00 \\
    WMD \cite{kusner2015word} & .67$\pm$.00 & .55$\pm$.00 & .61$\pm$.00 & .50$\pm$.00 & .60$\pm$.00 & .54$\pm$.00 & .66$\pm$.00 \\
    BERT \cite{devlin2018bert}&
    .79$\pm$.00 & .70$\pm$.00 & .74$\pm$.00 & .68$\pm$.00 & .59$\pm$.00 & .63$\pm$.00 & .64$\pm$.00 \\
    BiLSTM \cite{graves2005framewise} &
    .49$\pm$.00 & .43$\pm$.00 & .50$\pm$.00 & .49$\pm$.00 & .39$\pm$.00 & .46$\pm$.00 & .44$\pm$.00 \\
    PP-GCN \cite{peng2019fine} &.55$\pm$.02 & .51$\pm$.02 & .55$\pm$.02 & .50$\pm$.01 & .45$\pm$.01 & .47$\pm$.01 & .44$\pm$.01 \\
    EventX \cite{liu2020story}& .71$\pm$.00 & .67$\pm$.00 & .68$\pm$.00 & .65$\pm$.00 & .61$\pm$.00 & .58$\pm$.00 & .57$\pm$.00 \\
    \hline
    KPGNN$_t$ & \textit{.79$\pm$.01} & \textit{.74$\pm$.01} & \textit{.79$\pm$.01} & \textit{.73$\pm$.00} & \textbf{.69$\pm$.01} & \textit{.68$\pm$.01} & \textit{.68$\pm$.01} \\
    KPGNN& \textbf{.80$\pm$.00} & \textbf{.74$\pm$.02} & \textbf{.80$\pm$.01} & \textbf{.74$\pm$.01}& \textit{.68$\pm$.01} & \textbf{.69$\pm$.01} & \textbf{.69$\pm$.00} \\
    \toprule
    Blocks & $M_{15}$ & $M_{16}$ & $M_{17}$ & $M_{18}$ & $M_{19}$ & $M_{20}$ & $M_{21}$ \\ 
    \midrule
    Word2vec \cite{mikolov2013efficient} &.27$\pm$.00 & .49$\pm$.00 & .33$\pm$.00 & .29$\pm$.00 & .37$\pm$.00 & .38$\pm$.00 & .31$\pm$.00 \\
    LDA \cite{blei2003latent} & .21$\pm$.00 & .35$\pm$.01 & .19$\pm$.00 & .18$\pm$.00 & .29$\pm$.01 & .35$\pm$.00 & .19$\pm$.00 \\
    WMD \cite{kusner2015word}& .51$\pm$.00 & .60$\pm$.00 & .55$\pm$.00 & .63$\pm$.00 & .54$\pm$.00 & .58$\pm$.00 & .58$\pm$.00 \\
    BERT \cite{devlin2018bert}&
    .54$\pm$.00 & .75$\pm$.00 & .63$\pm$.00 & .57$\pm$.00 & .66$\pm$.00 & .68$\pm$.00 & .59$\pm$.00 \\
    BiLSTM \cite{graves2005framewise} &
    .40$\pm$.00 & .53$\pm$.00 & .45$\pm$.00 & .44$\pm$.00 & .44$\pm$.00 & .48$\pm$.00 & .41$\pm$.00 \\
    PP-GCN \cite{peng2019fine} & .39$\pm$.01 & .55$\pm$.01 & .48$\pm$.00 & .47$\pm$.01 & .51$\pm$.02 & .51$\pm$.01 & .41$\pm$.02 \\
    EventX \cite{liu2020story} & .49$\pm$.00 & .62$\pm$.00 & .58$\pm$.00 & .59$\pm$.00 & .60$\pm$.00 & .67$\pm$.00 & .53$\pm$.00 \\
    \hline
    KPGNN$_t$& \textit{.57$\pm$.01} & \textit{.78$\pm$.01} & \textit{.69$\pm$.01} & \textit{.68$\pm$.01} & \textit{.73$\pm$.00} & \textbf{.73$\pm$.00} & \textit{.59$\pm$.01} \\
    KPGNN& \textbf{.58$\pm$.00} & \textbf{.79$\pm$.01} & \textbf{.70$\pm$.01} & \textbf{.68$\pm$.02} & \textbf{.73$\pm$.01} & \textit{.72$\pm$.02} & \textbf{.60$\pm$.00} \\
    \bottomrule
  \end{tabular}
  \label{table:online_eval_nmi}
\end{table*}

\begin{table*}[t]
\addtolength{\tabcolsep}{+8pt}
\caption{Incremental evaluation AMIs. \textmd{The best results are marked in bold and second-best in italic.}}
  \centering
    \begin{tabular}{c|ccccccc}
   \toprule
    Blocks & $M_1$ & $M_2$ & $M_3$ & $M_4$ & $M_5$ & $M_6$ & $M_7$ \\
    \midrule
    Word2vec \cite{mikolov2013efficient} &.08$\pm$.00 & .41$\pm$.00 & .31$\pm$.00 & .24$\pm$.00 & .33$\pm$.00 & .40$\pm$.00 & .13$\pm$.00 \\
    LDA \cite{blei2003latent} & .08$\pm$.00 & .20$\pm$.01 & .22$\pm$.01 & .17$\pm$.00 & .21$\pm$.00 & .20$\pm$.00 & .12$\pm$.01 \\
    WMD \cite{kusner2015word}& .30$\pm$.00 & .69$\pm$.00 & .63$\pm$.00 & .45$\pm$.00 & .57$\pm$.00 & .57$\pm$.00 & .46$\pm$.00 \\
    BERT \cite{devlin2018bert}&
    .34$\pm$.00 & .76$\pm$.00 & .73$\pm$.00 & .55$\pm$.00 & \textit{.71$\pm$.00} & .74$\pm$.00 & .50$\pm$.00 \\
    BiLSTM \cite{graves2005framewise} &
    .12$\pm$.00 & .41$\pm$.00 & .31$\pm$.00 & .30$\pm$.00 & .33$\pm$.00 & .36$\pm$.00 & .20$\pm$.00 \\
    PP-GCN \cite{peng2019fine} &.21$\pm$.00 & .55$\pm$.02 & .52$\pm$.01 & .42$\pm$.01 & .46$\pm$.01 & .52$\pm$.02 & .34$\pm$.00 \\
    EventX \cite{liu2020story}& .06$\pm$.00 & .29$\pm$.00 & .18$\pm$.00 & .19$\pm$.00 & .14$\pm$.00 & .27$\pm$.00 & .13$\pm$.00 \\
    \hline
    KPGNN$_t$&\textit{.36$\pm$.01} & \textit{.77$\pm$.01} & \textbf{.75$\pm$.00} & \textbf{.65$\pm$.01} & \textbf{.71$\pm$.01} & \textit{.78$\pm$.00} & \textit{.50$\pm$.01} \\
    KPGNN &\textbf{.37$\pm$.00} & \textbf{.78$\pm$.01} & \textit{.74$\pm$.00} & \textit{.64$\pm$.01} & \textbf{.71$\pm$.01} & \textbf{.79$\pm$.01} & \textbf{.51$\pm$.01} \\
    \toprule
    Blocks & $M_8$ & $M_9$ & $M_{10}$ & $M_{11}$& $M_{12}$ & $M_{13}$ & $M_{14}$ \\
    \midrule
    Word2vec \cite{mikolov2013efficient} &.33$\pm$.00 & .24$\pm$.00 & .39$\pm$.00 & .26$\pm$.00 & .23$\pm$.00 & .23$\pm$.00 & .26$\pm$.00 \\
    LDA \cite{blei2003latent} & 24$\pm$.01 & .24$\pm$.00 & .36$\pm$.01 & .25$\pm$.01 & .16$\pm$.01 & .19$\pm$.00 & .15$\pm$.00 \\
    WMD \cite{kusner2015word} & .63$\pm$.00 & .46$\pm$.00 & .57$\pm$.00 & .42$\pm$.00 & .58$\pm$.00 & .50$\pm$.00 & .64$\pm$.00 \\
    BERT \cite{devlin2018bert}&
    .75$\pm$.00 & .66$\pm$.00 & .70$\pm$.00 & .65$\pm$.00 & .56$\pm$.00 & .59$\pm$.00 & .61$\pm$.00 \\
    BiLSTM \cite{graves2005framewise} &
    .35$\pm$.00 & .32$\pm$.00 & .39$\pm$.00 & .37$\pm$.00 & .32$\pm$.00 & .31$\pm$.00 & .34$\pm$.00 \\
    PP-GCN \cite{peng2019fine} &.49$\pm$.02 & .46$\pm$.02 & .51$\pm$.02 & .46$\pm$.01 & .42$\pm$.01 & .43$\pm$.01 & .41$\pm$.01 \\
    EventX \cite{liu2020story}& .21$\pm$.00 & .19$\pm$.00 & .24$\pm$.00 & .24$\pm$.00 & .16$\pm$.00 & .16$\pm$.00 & .14$\pm$.00 \\
    \hline
    KPGNN$_t$ & \textit{.75$\pm$.01} & \textit{.70$\pm$.01} & \textit{.76$\pm$.01} & \textit{.70$\pm$.00} & \textbf{.66$\pm$.01} & \textit{.65$\pm$.01} & \textbf{.65$\pm$.01} \\
    KPGNN& \textbf{.76$\pm$.01} & \textbf{.71$\pm$.02} & \textbf{.78$\pm$.01} & \textbf{.71$\pm$.01} & \textbf{.66$\pm$.01} & \textbf{.67$\pm$.01} & \textit{.65$\pm$.00} \\
    \toprule
    Blocks & $M_{15}$ & $M_{16}$ & $M_{17}$ & $M_{18}$ & $M_{19}$ & $M_{20}$ & $M_{21}$ \\ 
    \midrule
    Word2vec \cite{mikolov2013efficient} &.15$\pm$.00 & .36$\pm$.00 & .24$\pm$.00 & .21$\pm$.00 & .28$\pm$.00 & .24$\pm$.00 & .21$\pm$.00 \\
    LDA \cite{blei2003latent} & .13$\pm$.00 & .27$\pm$.01 & .13$\pm$.00 & .12$\pm$.00 & .22$\pm$.01 & .23$\pm$.00 & .13$\pm$.00 \\
    WMD \cite{kusner2015word}& .47$\pm$.00 & .59$\pm$.00 & .57$\pm$.00 & .60$\pm$.00 & .49$\pm$.00 & .55$\pm$.00 & .52$\pm$.00 \\ 
    BERT \cite{devlin2018bert}&
    .50$\pm$.00 & .72$\pm$.00 & .60$\pm$.00 & .53$\pm$.00 & .63$\pm$.00 & .62$\pm$.00 & \textit{.57$\pm$.00} \\
    BiLSTM \cite{graves2005framewise} &
    .26$\pm$.00 & .41$\pm$.00 & .35$\pm$.00 & .35$\pm$.00 & .35$\pm$.00 & .34$\pm$.00 & .31$\pm$.00 \\
    PP-GCN \cite{peng2019fine} & .35$\pm$.01 & .52$\pm$.01 & .45$\pm$.00 & .45$\pm$.01 & .48$\pm$.02 & .45$\pm$.02 & .38$\pm$.02 \\
    EventX \cite{liu2020story} & .07$\pm$.00 & .19$\pm$.00 & .18$\pm$.00 & .16$\pm$.00 & .16$\pm$.00 & .18$\pm$.00 & .10$\pm$.00 \\
    \hline
    KPGNN$_t$& \textit{.53$\pm$.01} & \textit{.75$\pm$.01} & \textit{.67$\pm$.01} & \textit{.66$\pm$.01} & \textit{.70$\pm$.00} & \textit{.68$\pm$.00} & \textbf{.57$\pm$.01} \\
    KPGNN& \textbf{.54$\pm$.00} & \textbf{.77$\pm$.01} & \textbf{.68$\pm$.01} & \textbf{.66$\pm$.02} & \textbf{.71$\pm$.01} & \textbf{.68$\pm$.02} & \textit{.57$\pm$.00} \\
    \bottomrule
  \end{tabular}
  \label{table:online_eval_ami}
\end{table*}

\begin{table*}[t]
\addtolength{\tabcolsep}{+8pt}
\caption{Incremental evaluation ARIs. \textmd{The best results are marked in bold and second-best in italic.}}
  \centering
    \begin{tabular}{c|ccccccc}
   \toprule
    Blocks & $M_1$ & $M_2$ & $M_3$ & $M_4$ & $M_5$ & $M_6$ & $M_7$ \\
    \midrule
    Word2vec \cite{mikolov2013efficient} &01$\pm$.00 & .49$\pm$.00 & .16$\pm$.00 & .07$\pm$.00 & .17$\pm$.00 & .25$\pm$.00 & .02$\pm$.00 \\
    LDA \cite{blei2003latent} & .00$\pm$.00 & .08$\pm$.00 & .02$\pm$.01 & .07$\pm$.00 & .06$\pm$.00 & .07$\pm$.01 & .00$\pm$.00 \\
    WMD \cite{kusner2015word}& .04$\pm$.00 & .48$\pm$.00 & .28$\pm$.00 & .11$\pm$.00 & .26$\pm$.00 & .16$\pm$.00 & .08$\pm$.00 \\
    BERT \cite{devlin2018bert}&
    03$\pm$.00 & .64$\pm$.00 & .43$\pm$.00 & .19$\pm$.00 & .44$\pm$.00 & .44$\pm$.00 & .07$\pm$.00 \\
    BiLSTM \cite{graves2005framewise} &
    .03$\pm$.00 & .49$\pm$.00 & .17$\pm$.00 & .11$\pm$.00 & .19$\pm$.00 & .18$\pm$.00 & \textbf{.12$\pm$.00} \\
    PP-GCN \cite{peng2019fine} &.05$\pm$.00 & .67$\pm$.03 & .47$\pm$.01 & .24$\pm$.01 & .34$\pm$.00 & .55$\pm$.03 & \textit{.11$\pm$.02} \\
    EventX \cite{liu2020story}& .01$\pm$.00 & .45$\pm$.00 & .09$\pm$.00 & .07$\pm$.00 & .04$\pm$.00 & .14$\pm$.00 & .02$\pm$.00 \\
    \hline
    KPGNN$_t$&\textit{.06$\pm$.01} & \textit{.76$\pm$.01} & \textbf{.60$\pm$.02} & \textbf{.30$\pm$.01} & \textbf{.48$\pm$.01} & \textit{.67$\pm$.05} & .11$\pm$.01 \\
    KPGNN &\textbf{.07$\pm$.01} & \textbf{.76$\pm$.02} & \textit{.58$\pm$.01} & \textit{.29$\pm$.01} & \textit{.47$\pm$.03} & \textbf{.72$\pm$.03} & \textbf{.12$\pm$.00} \\
    \toprule
    Blocks & $M_8$ & $M_9$ & $M_{10}$ & $M_{11}$& $M_{12}$ & $M_{13}$ & $M_{14}$ \\
    \midrule
    Word2vec \cite{mikolov2013efficient} &.17$\pm$.00 & .08$\pm$.00 & .23$\pm$.00 & .09$\pm$.00 & .09$\pm$.00 & .06$\pm$.00 & .10$\pm$.00 \\
    LDA \cite{blei2003latent} & .03$\pm$.00 & .03$\pm$.01 & .09$\pm$.02 & .03$\pm$.01 & .02$\pm$.00 & .00$\pm$.00 & .02$\pm$.00 \\
    WMD \cite{kusner2015word} & .22$\pm$.00 & .12$\pm$.00 & .20$\pm$.00 & .12$\pm$.00 & .27$\pm$.00 & .13$\pm$.00 & .33$\pm$.00 \\
    BERT \cite{devlin2018bert}&
    .50$\pm$.00 & .33$\pm$.00 & .44$\pm$.00 & .27$\pm$.00 & .31$\pm$.00 & .14$\pm$.00 & .30$\pm$.00 \\
    BiLSTM \cite{graves2005framewise} &
    .17$\pm$.00 & .13$\pm$.00 & .30$\pm$.00 & .16$\pm$.00 & .14$\pm$.00 & .10$\pm$.00 & .17$\pm$.00 \\
    PP-GCN \cite{peng2019fine} &.43$\pm$.04 & .31$\pm$.02 & .50$\pm$.07 & .38$\pm$.02 & .34$\pm$.03 & .19$\pm$.01 & .29$\pm$.01 \\
    EventX \cite{liu2020story}& .09$\pm$.00 & .07$\pm$.00 & .13$\pm$.00 & .16$\pm$.00 & .07$\pm$.00 & .04$\pm$.00 & .10$\pm$.00 \\
    \hline
    KPGNN$_t$& \textit{.59$\pm$.02} & \textit{.45$\pm$.02} & \textit{.64$\pm$.01} & \textit{.48$\pm$.01} & \textbf{.50$\pm$.03} & \textit{.28$\pm$.01} & \textbf{.43$\pm$.02} \\
    KPGNN& \textbf{.60$\pm$.01} & \textbf{.46$\pm$.02} & \textbf{.70$\pm$.06} & \textbf{.49$\pm$.03} & \textit{.48$\pm$.01} & \textbf{.29$\pm$.03} & \textit{.42$\pm$.02} \\
    \toprule
    Blocks & $M_{15}$ & $M_{16}$ & $M_{17}$ & $M_{18}$ & $M_{19}$ & $M_{20}$ & $M_{21}$ \\ 
    \midrule
    Word2vec \cite{mikolov2013efficient} &.03$\pm$.00 & .19$\pm$.00 & .10$\pm$.00 & .07$\pm$.00 & .14$\pm$.00 & .10$\pm$.00 & .06$\pm$.00 \\
    LDA \cite{blei2003latent} & .00$\pm$.00 & .11$\pm$.01 & .02$\pm$.00 & .02$\pm$.00 & .03$\pm$.00 & .02$\pm$.01 & .00$\pm$.01 \\
    WMD \cite{kusner2015word}& .16$\pm$.00 & .32$\pm$.00 & .26$\pm$.00 & .35$\pm$.00 & .12$\pm$.00 & .19$\pm$.00 & .19$\pm$.00 \\
    BERT \cite{devlin2018bert}&
    .10$\pm$.00 & .41$\pm$.00 & .24$\pm$.00 & .24$\pm$.00 & .32$\pm$.00 & .33$\pm$.00 & .18$\pm$.00 \\
    BiLSTM \cite{graves2005framewise} &
    .08$\pm$.00 & .27$\pm$.00 & .22$\pm$.00 & .19$\pm$.00 & .16$\pm$.00 & .20$\pm$.00 & .16$\pm$.00 \\
    PP-GCN \cite{peng2019fine} & .15$\pm$.00 & .51$\pm$.03 & .35$\pm$.03 & .39$\pm$.03 & .41$\pm$.02 & .41$\pm$.01 & \textit{.20$\pm$.03} \\
    EventX \cite{liu2020story} & .01$\pm$.00 & .08$\pm$.00 & .12$\pm$.00 & .08$\pm$.00 & .07$\pm$.00 & .11$\pm$.00 & .01$\pm$.00 \\
    \hline
    KPGNN$_t$& \textit{.16$\pm$.02} & \textit{.62$\pm$.03} & \textit{.41$\pm$.03} & \textit{.46$\pm$.02} & \textit{.50$\pm$.01} & \textit{.51$\pm$.01} & \textbf{.23$\pm$.02} \\
    KPGNN& \textbf{.17$\pm$.00} & \textbf{.66$\pm$.05} & \textbf{.43$\pm$.05} & \textbf{.47$\pm$.04} & \textbf{.51$\pm$.03} & \textbf{.51$\pm$.04} & .20$\pm$.01 \\
    \bottomrule
  \end{tabular}
  \label{table:online_eval_ARI}
\end{table*}

\begin{figure*}[h]
    \centering
    {
    \begin{minipage}[c]{0.3\textwidth}
    \centering
        \includegraphics[width =5cm]{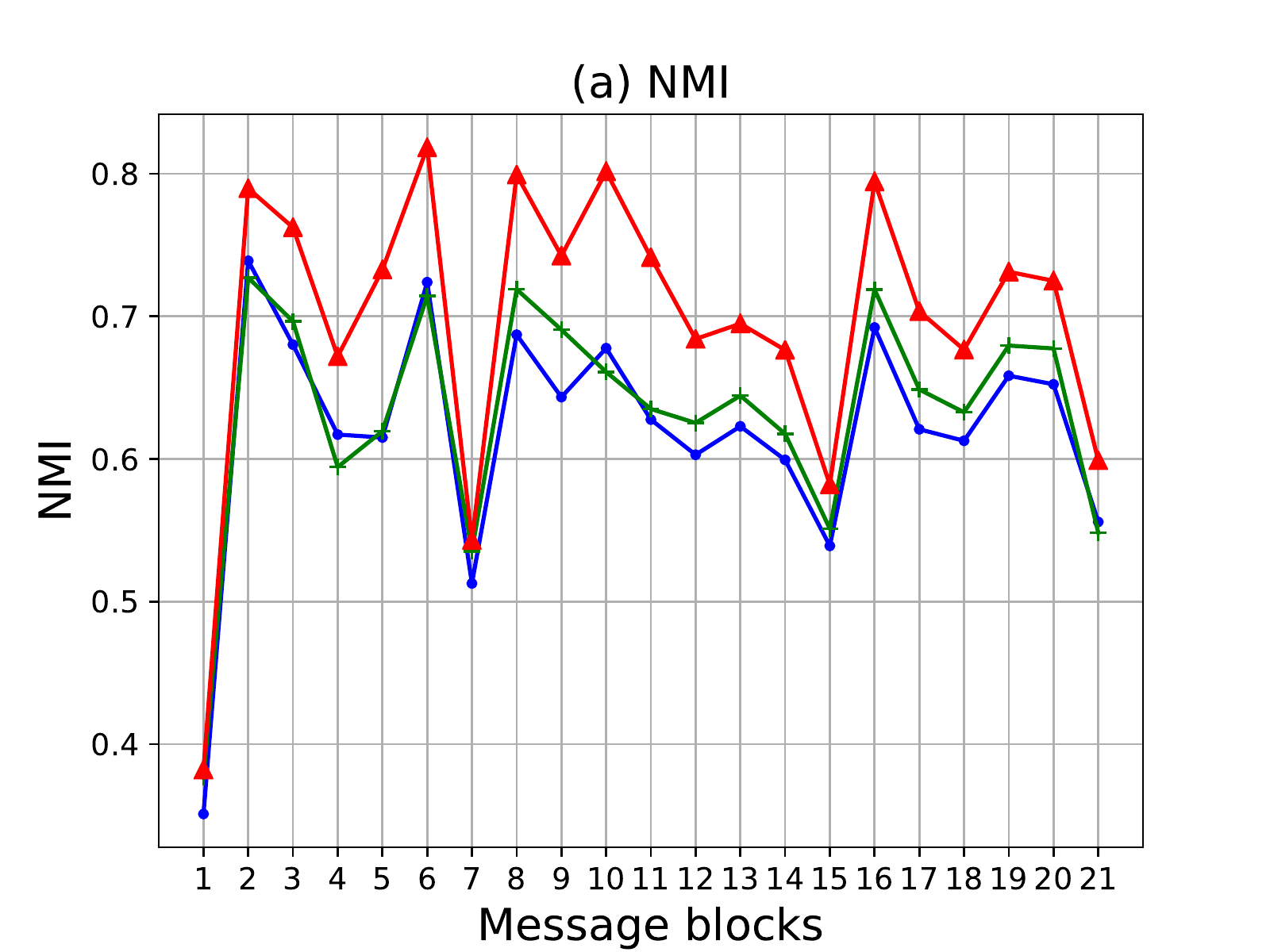}
    \end{minipage}
    }
    \centering
    {
    \begin{minipage}[c]{0.3\textwidth}
    \centering
        \includegraphics[width =5cm]{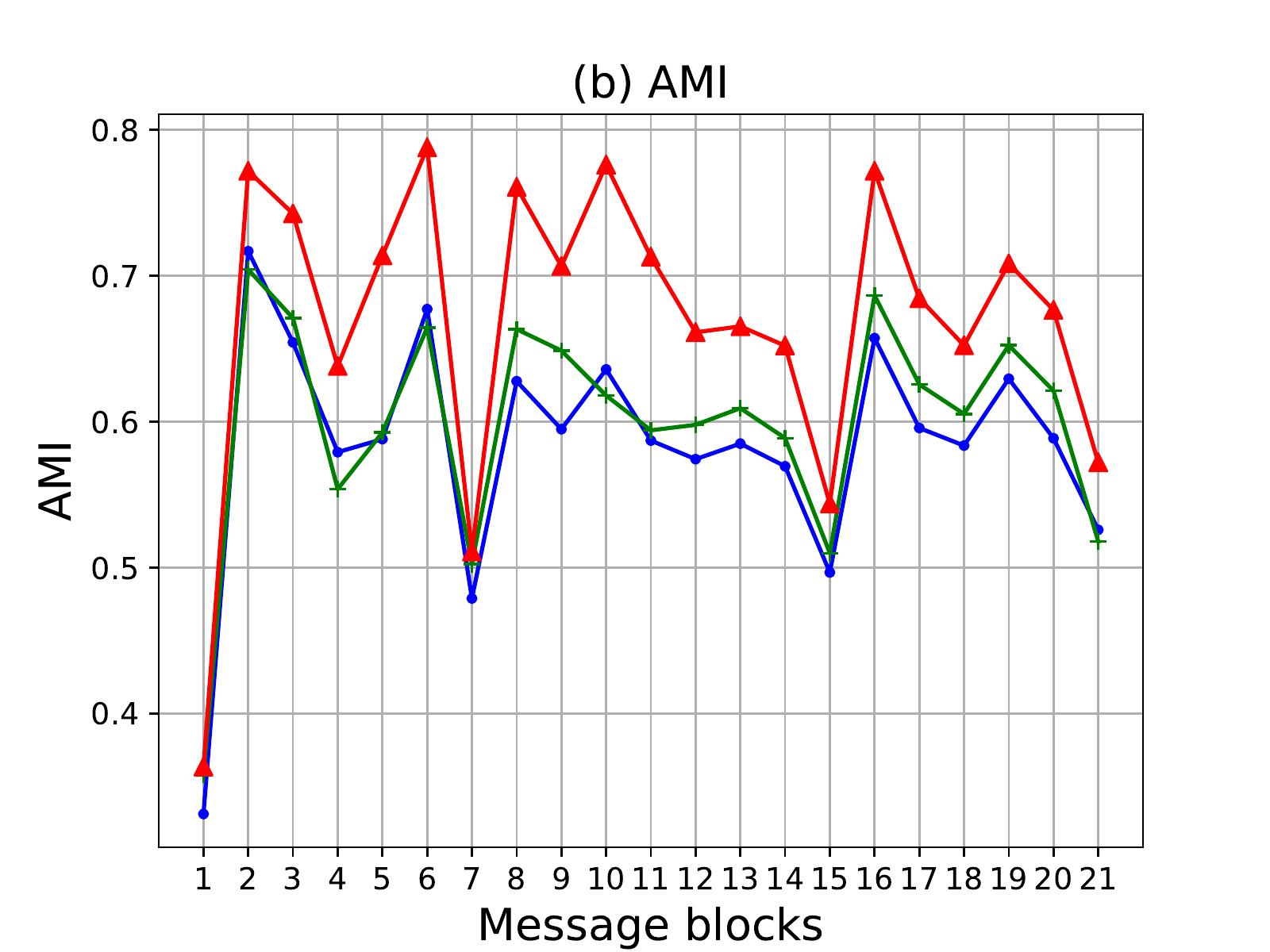}
    \end{minipage}
    }
    \centering
    {
    \begin{minipage}[c]{0.3\textwidth}
    \centering
        \includegraphics[width =5cm]{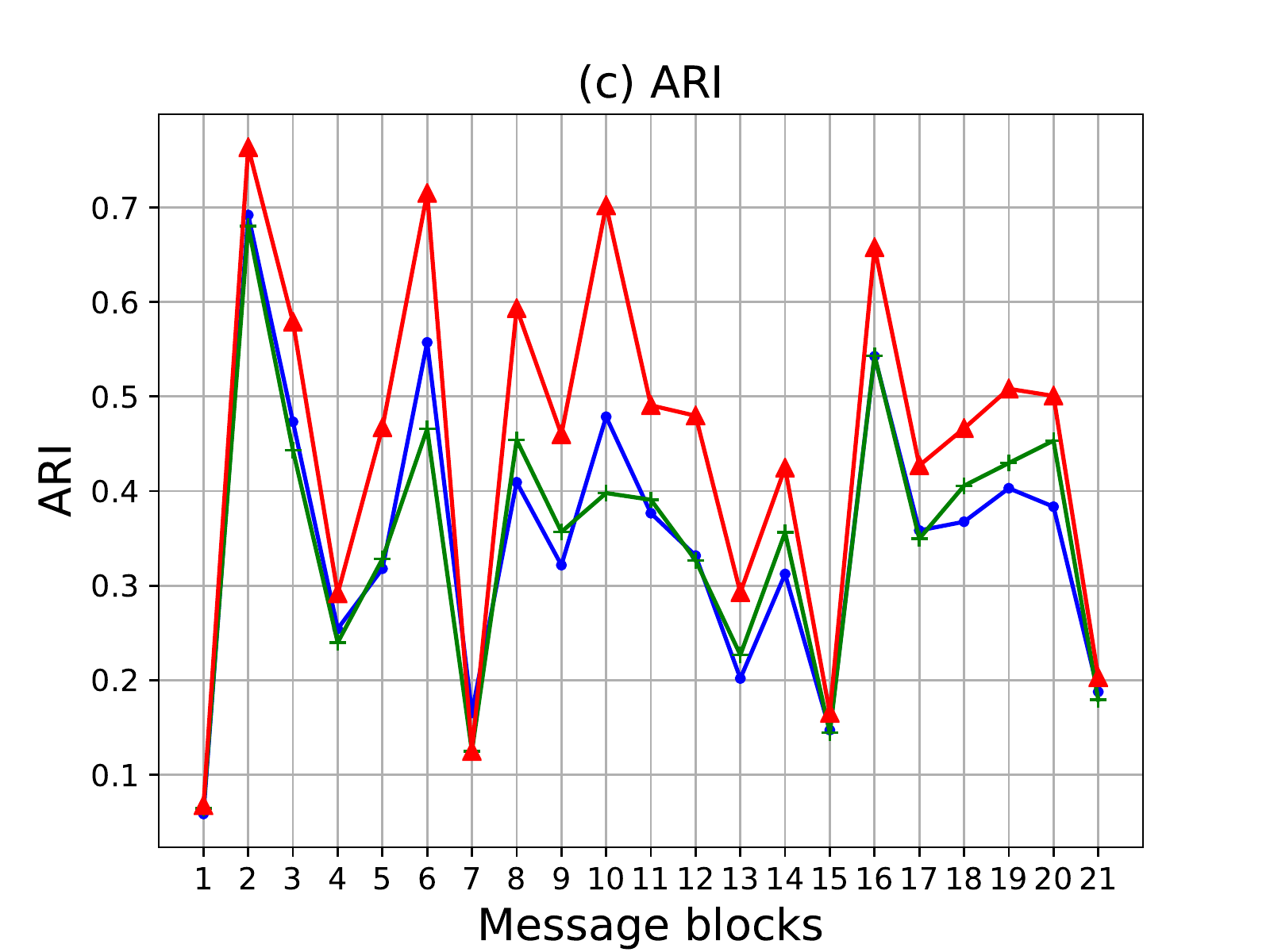}
    \end{minipage}
    }
    \centering
    {
    \begin{minipage}[c]{0.3\textwidth}
    \centering
        \includegraphics[width =5cm]{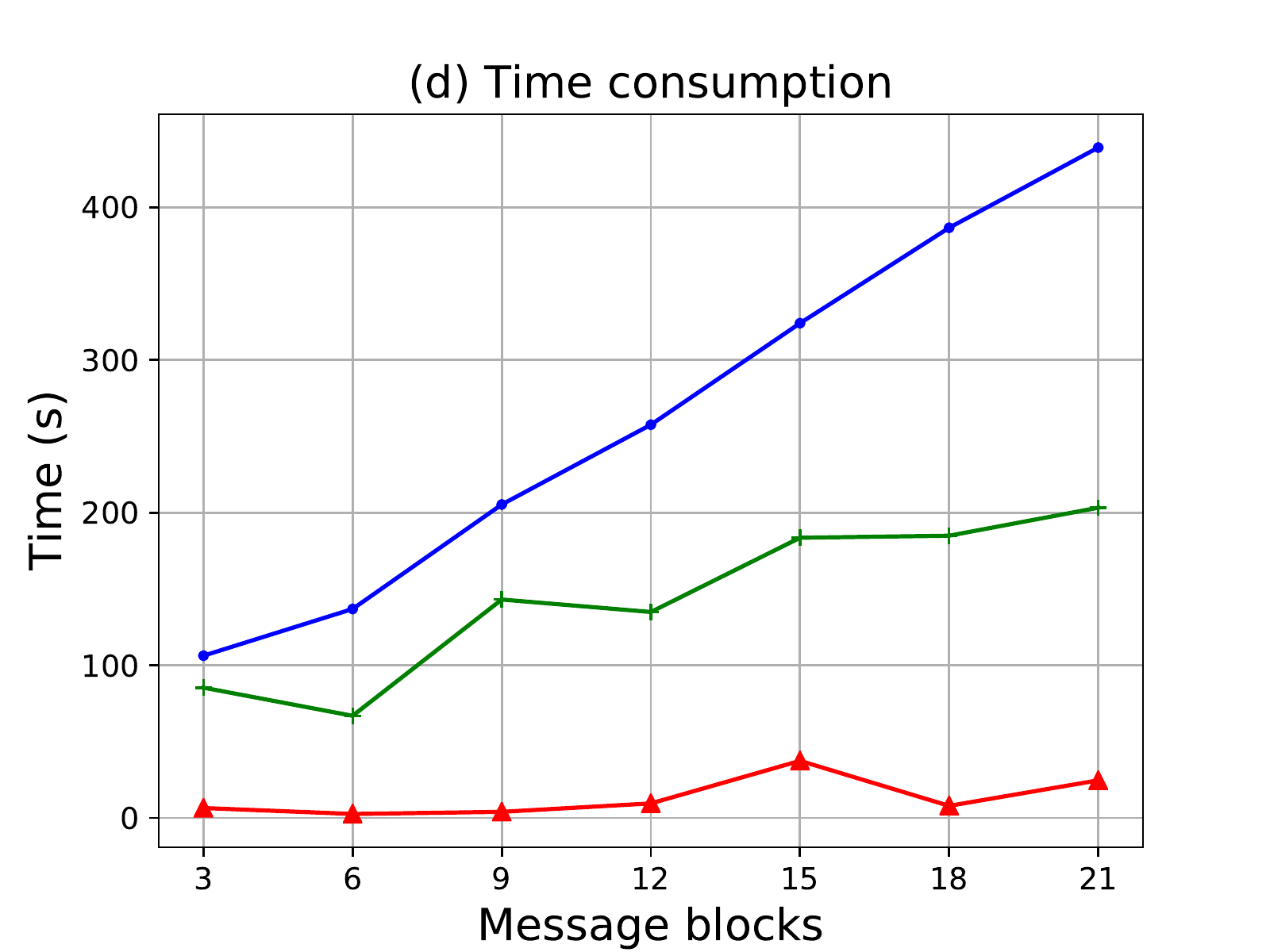}
    \end{minipage}
    }
    \centering
    {
    \begin{minipage}[c]{0.3\textwidth}
    \centering
        \includegraphics[width =5cm]{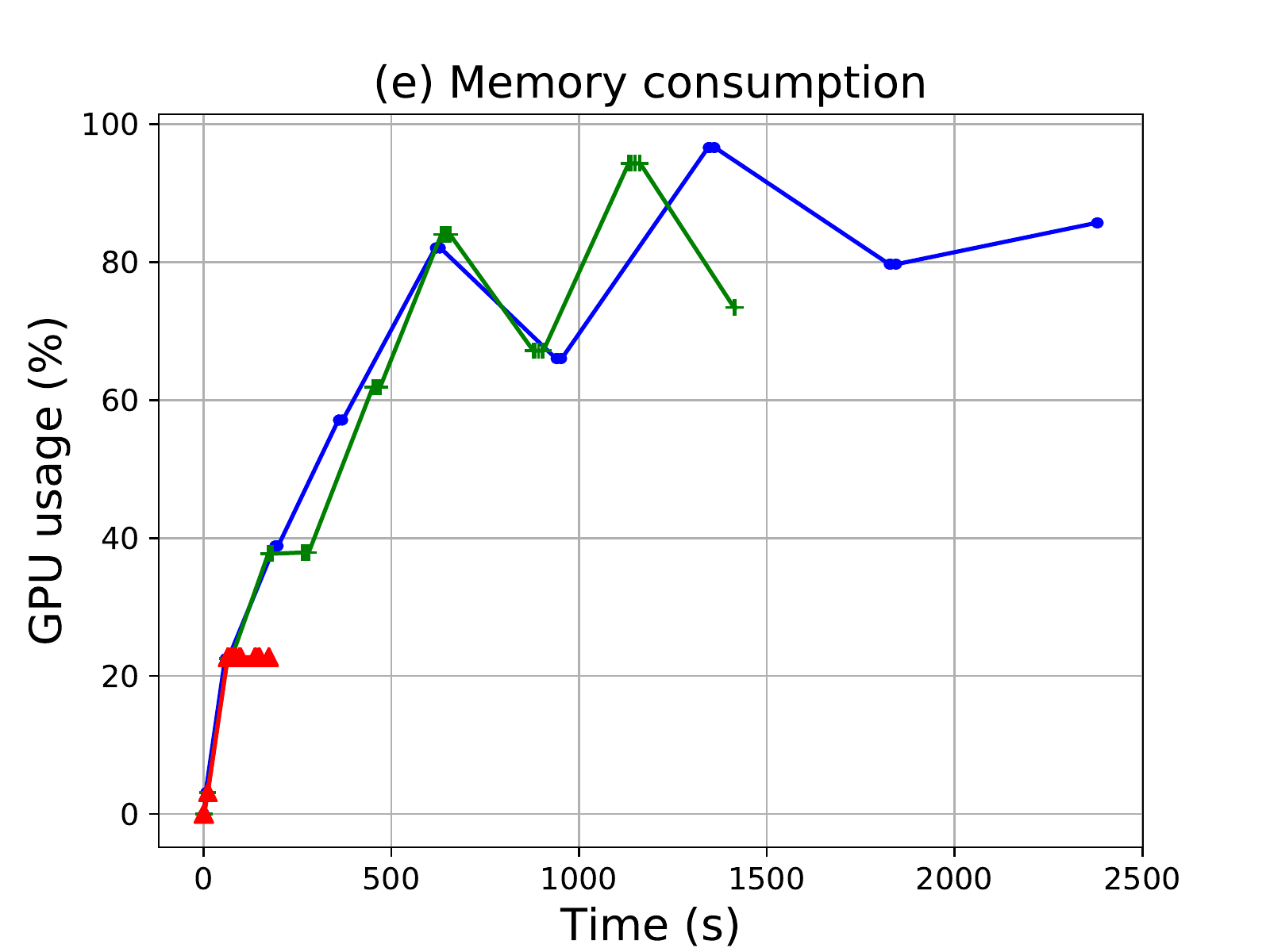}
    \end{minipage}
    }
    \centering
    {
    \begin{minipage}[c]{0.13\textwidth}
        \includegraphics[width =2cm]{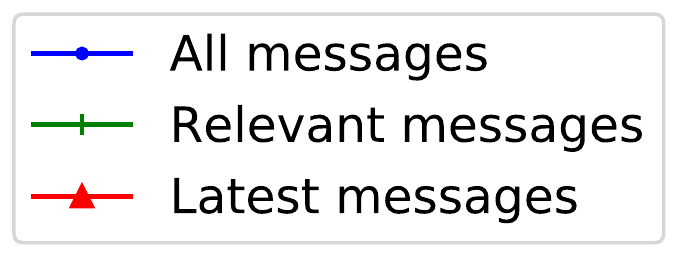}
    \end{minipage}
    }
    \caption{KPGNN with different update-maintenance strategies. \textmd{\textbf{(a)}, \textbf{(b)}, and \textbf{(c)} show the NMI, AMI, and ARI performance of KPGNN when adopting different update-maintenance strategies. In \textbf{(d)} and \textbf{(e)}, we train KPGNN for one mini-batch in the maintenance stages and measure time and memory consumption. \textbf{(d)} shows the time (in seconds) used for training KPGNN for one mini-batch. \textbf{(e)} shows the GPU$\%$ used over time throughout the training.}}
    \label{fig:strategy}
    \Description[The latest message strategy outperforms the others in terms of performance, time consumption, and memory consumption]{Sub-graphs a, b, and c show NMI, AMI, and ARI on the Y axes against message blocks from 1 to 21 on the X axes. Three lines are shown in each of them. The latest message strategy, the relevant message strategy, and the all message strategy are consistently the highest, second highest, and lowest. Sub-graph d shows time on the Y axis against message blocks from 1 to 21 on the X axis. The latest message strategy, the relevant message strategy, and the all message strategy are consistently the lowest, second lowest, and highest. Sub-graph e shows percentages of GPU usage on the Y axis against time on the X axis. The latest message strategy takes the least amounts of memory and time, the relevant message strategy the second least, and the all message strategy the highest.}
\end{figure*}

\begin{figure*}[h]
    \centering
    {
    \begin{minipage}[c]{0.24\textwidth}
    \centering
        \includegraphics[width =4.3cm]{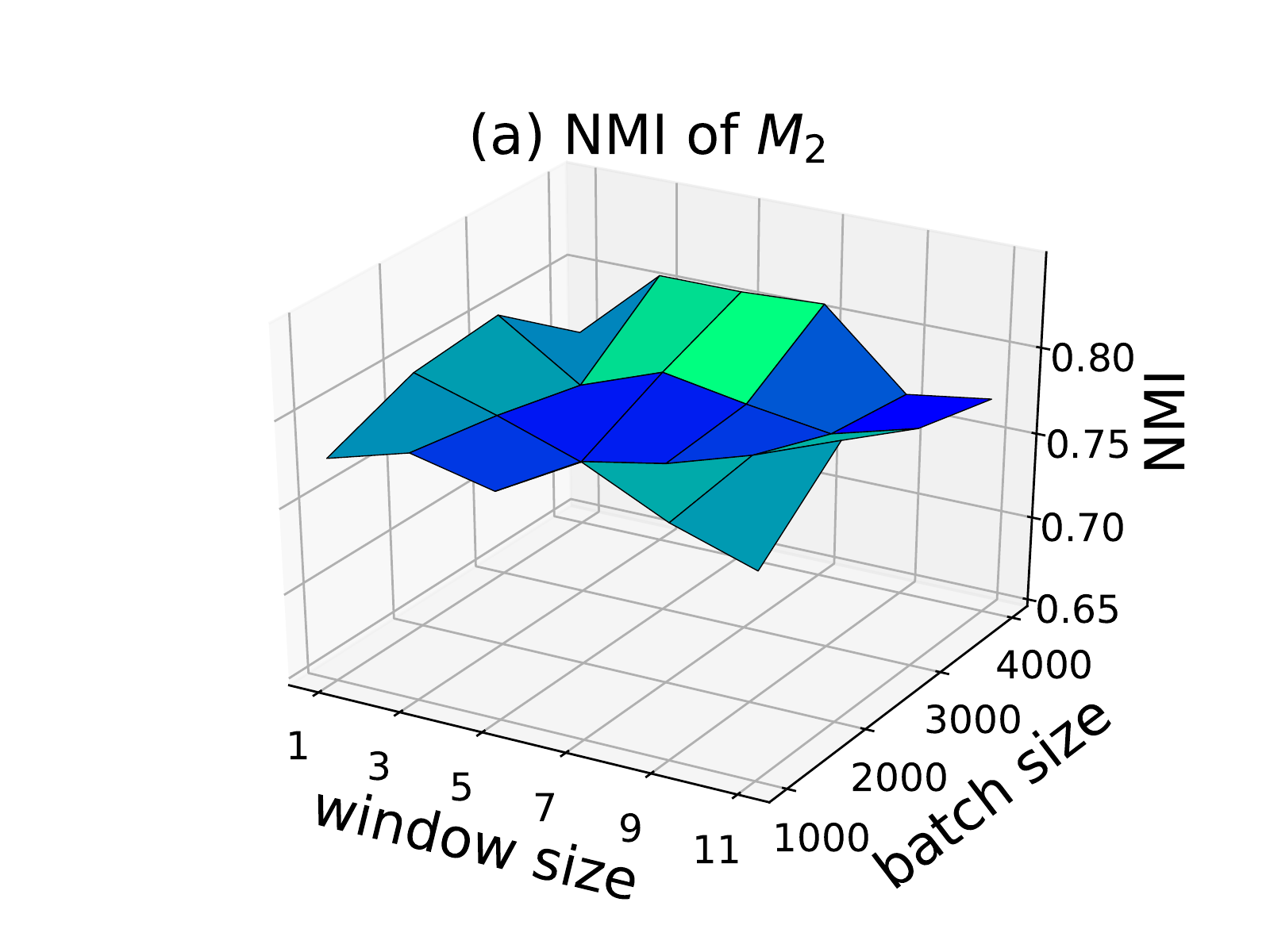}
    \end{minipage}
    }
    \centering
    {
    \begin{minipage}[c]{0.24\textwidth}
    \centering
        \includegraphics[width =4.3cm]{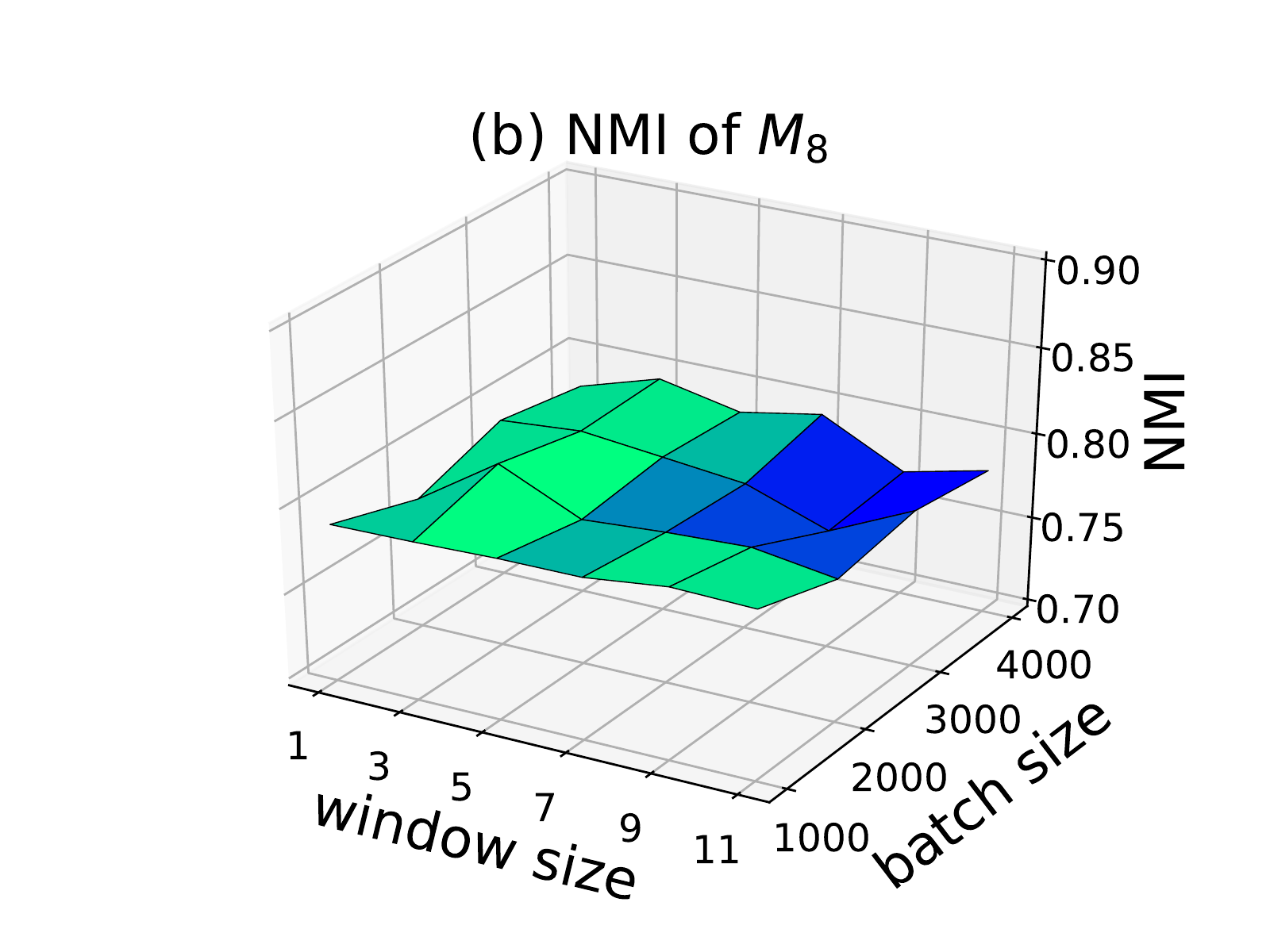}
    \end{minipage}
    }
    \centering
    {
    \begin{minipage}[c]{0.24\textwidth}
    \centering
        \includegraphics[width =4.3cm]{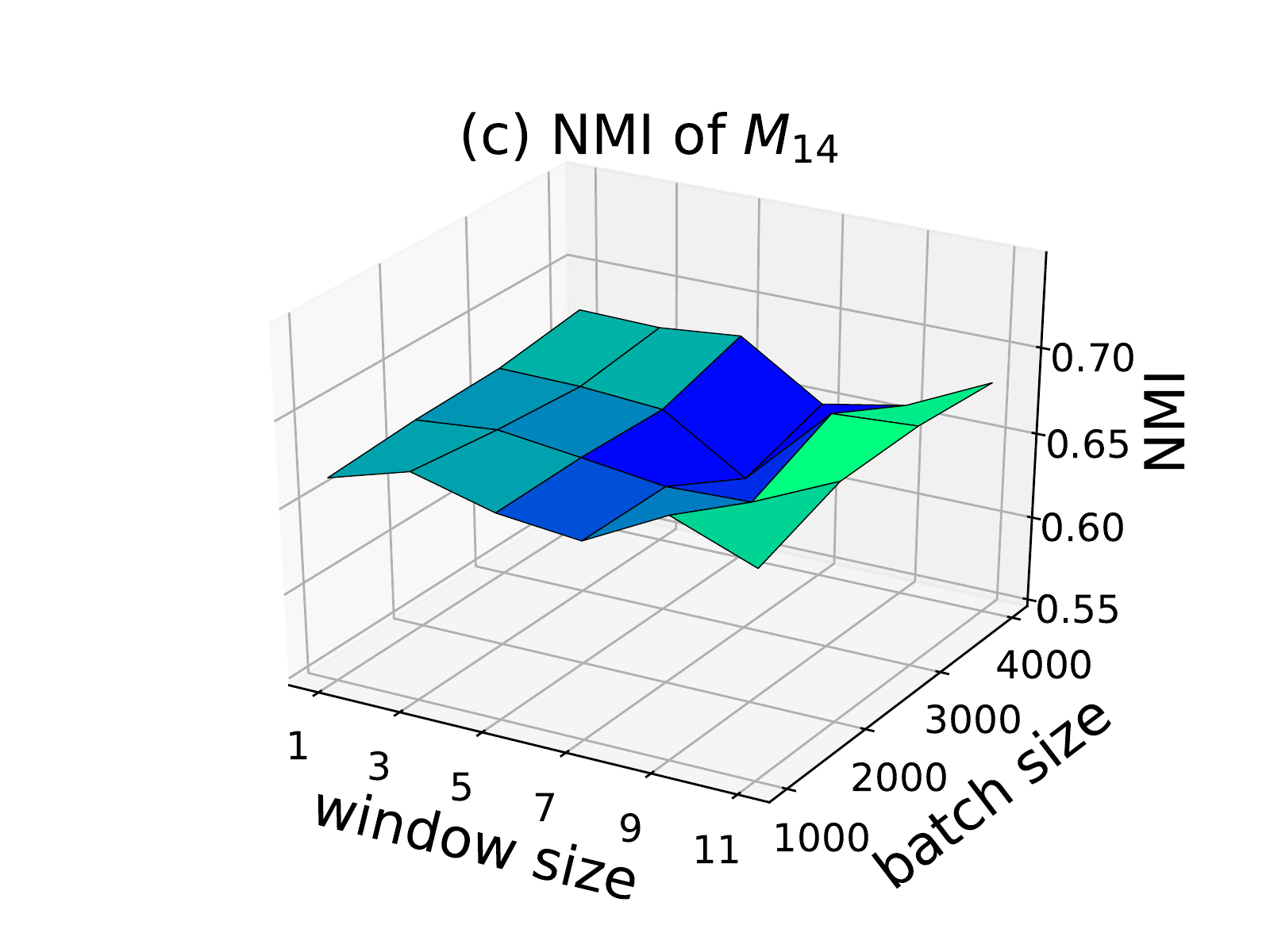}
    \end{minipage}
    }
    {
    \begin{minipage}[c]{0.24\textwidth}
    \centering
        \includegraphics[width =4.3cm]{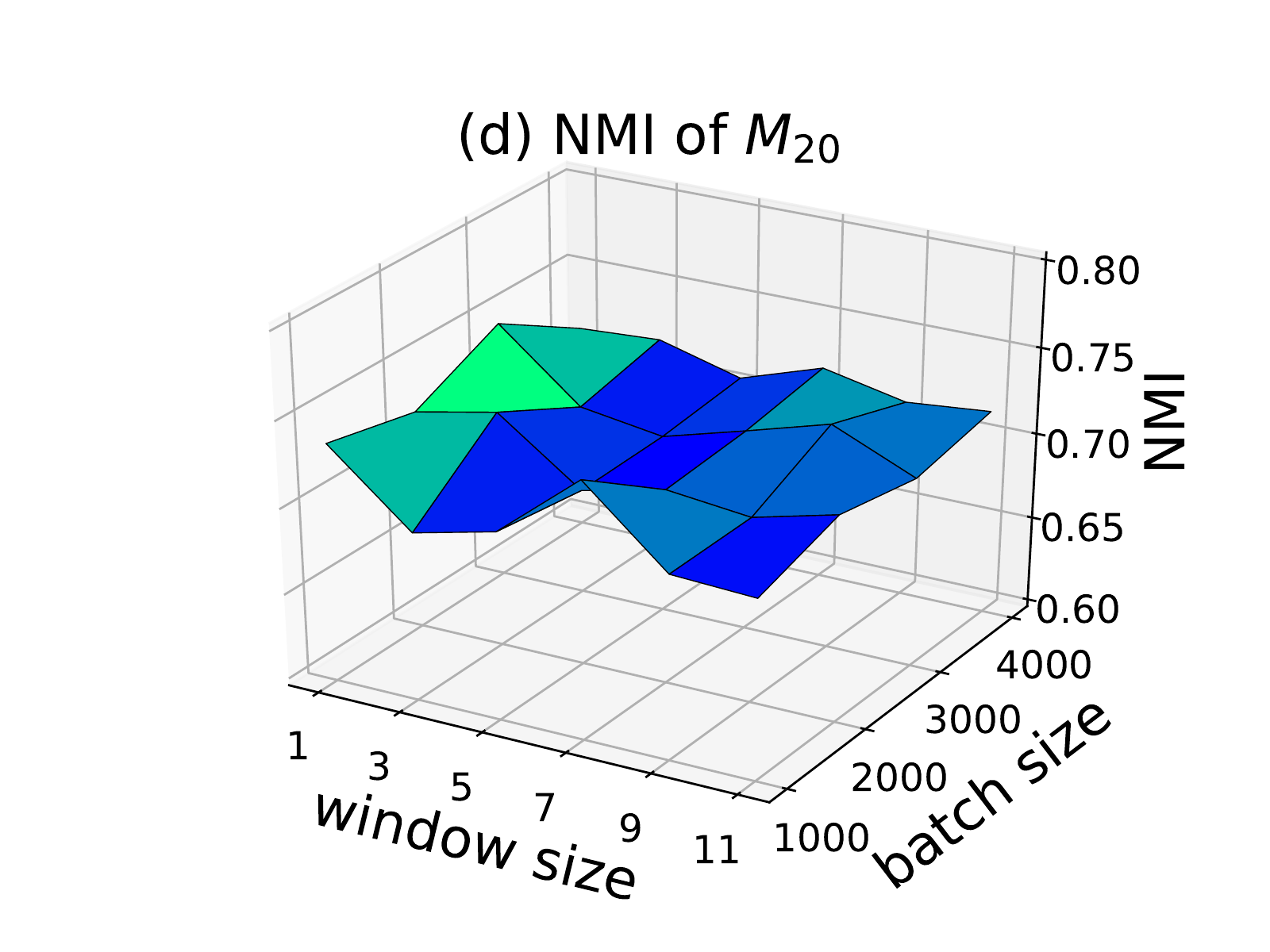}
    \end{minipage}
    }
    \centering
    {
    \begin{minipage}[c]{0.24\textwidth}
    \centering
        \includegraphics[width =4.3cm]{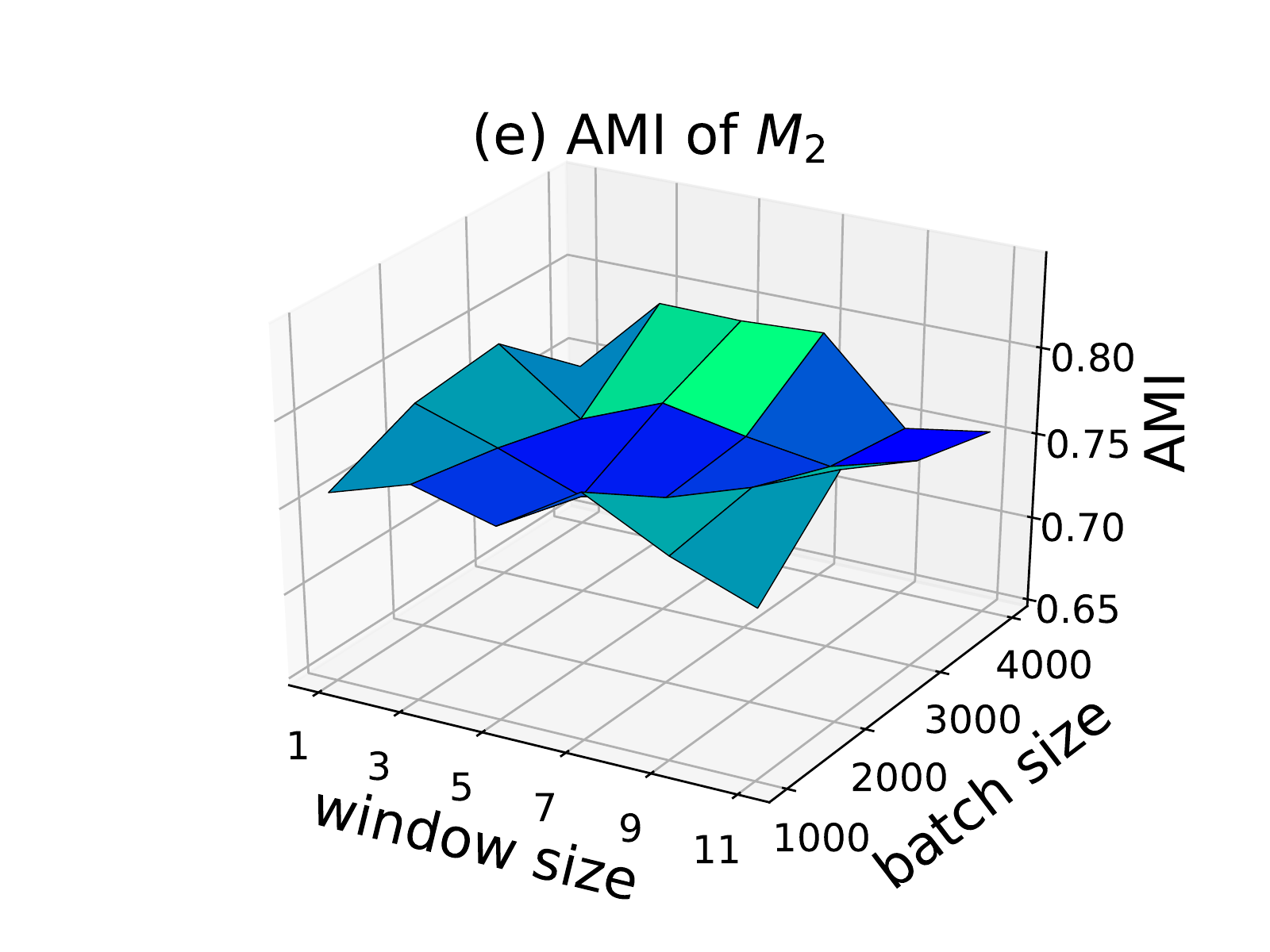}
    \end{minipage}
    }
    \centering
    {
    \begin{minipage}[c]{0.24\textwidth}
    \centering
        \includegraphics[width =4.3cm]{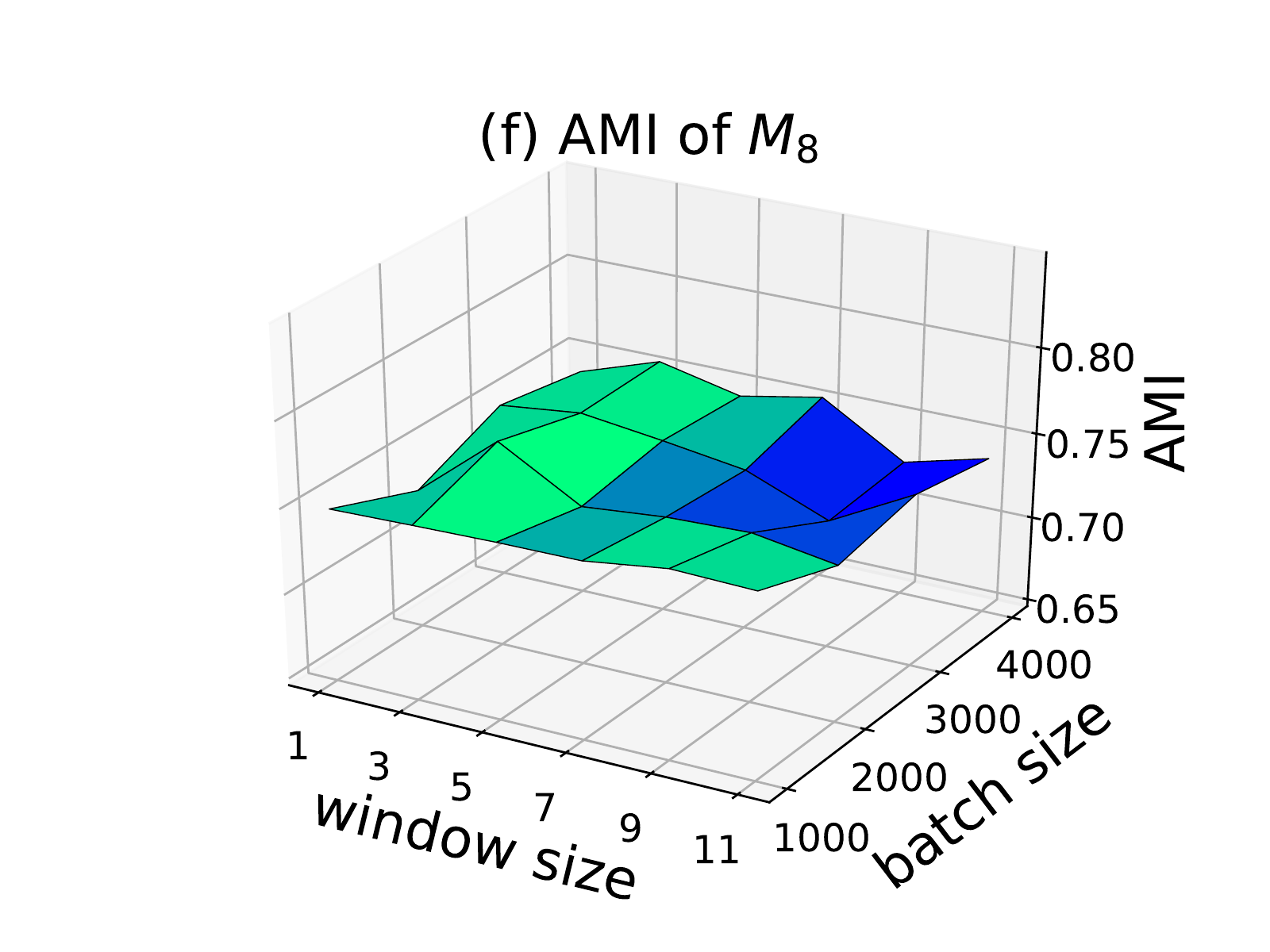}
    \end{minipage}
    }
    \centering
    {
    \begin{minipage}[c]{0.24\textwidth}
    \centering
        \includegraphics[width =4.3cm]{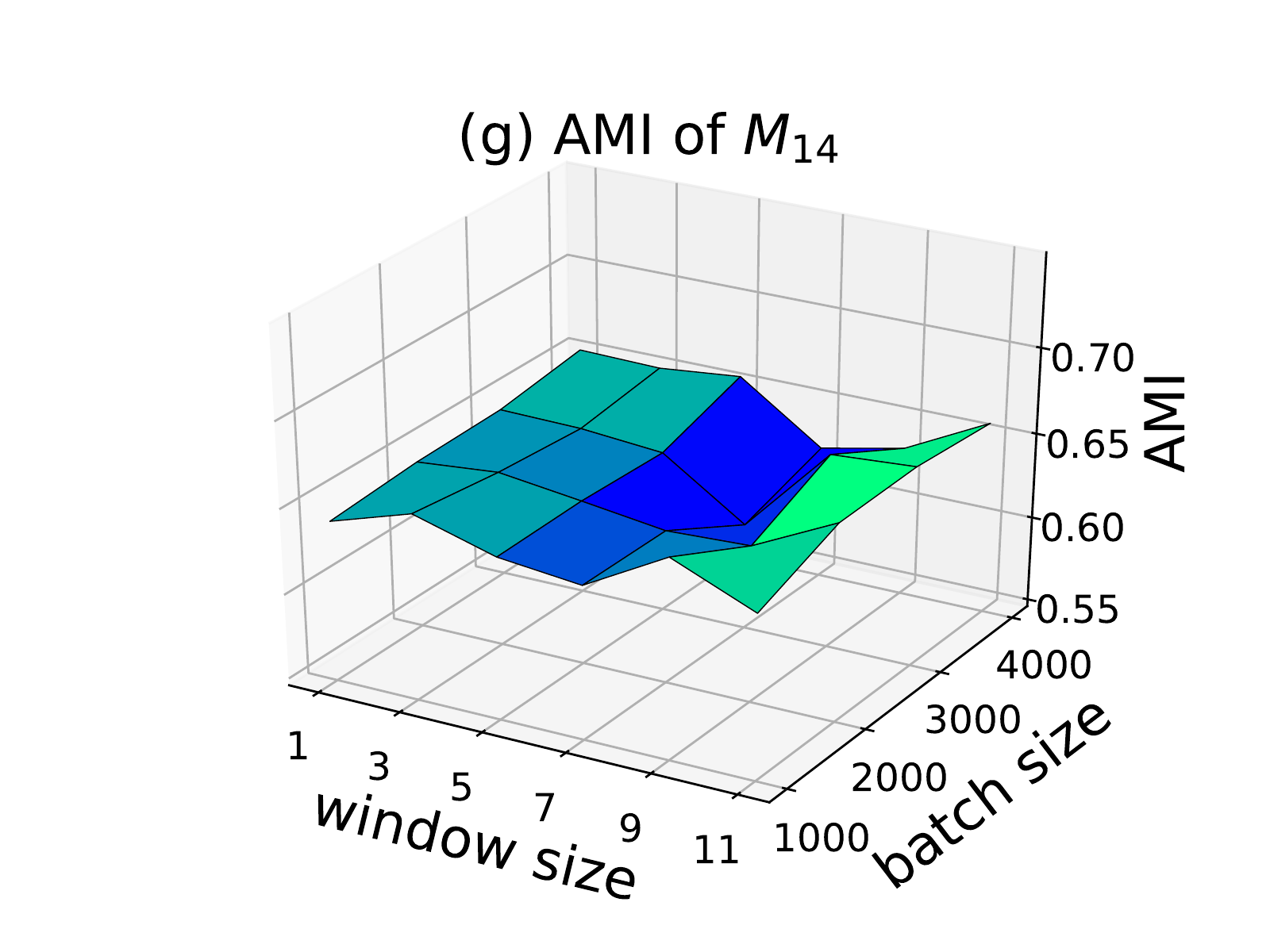}
    \end{minipage}
    }
    {
    \begin{minipage}[c]{0.24\textwidth}
    \centering
        \includegraphics[width =4.3cm]{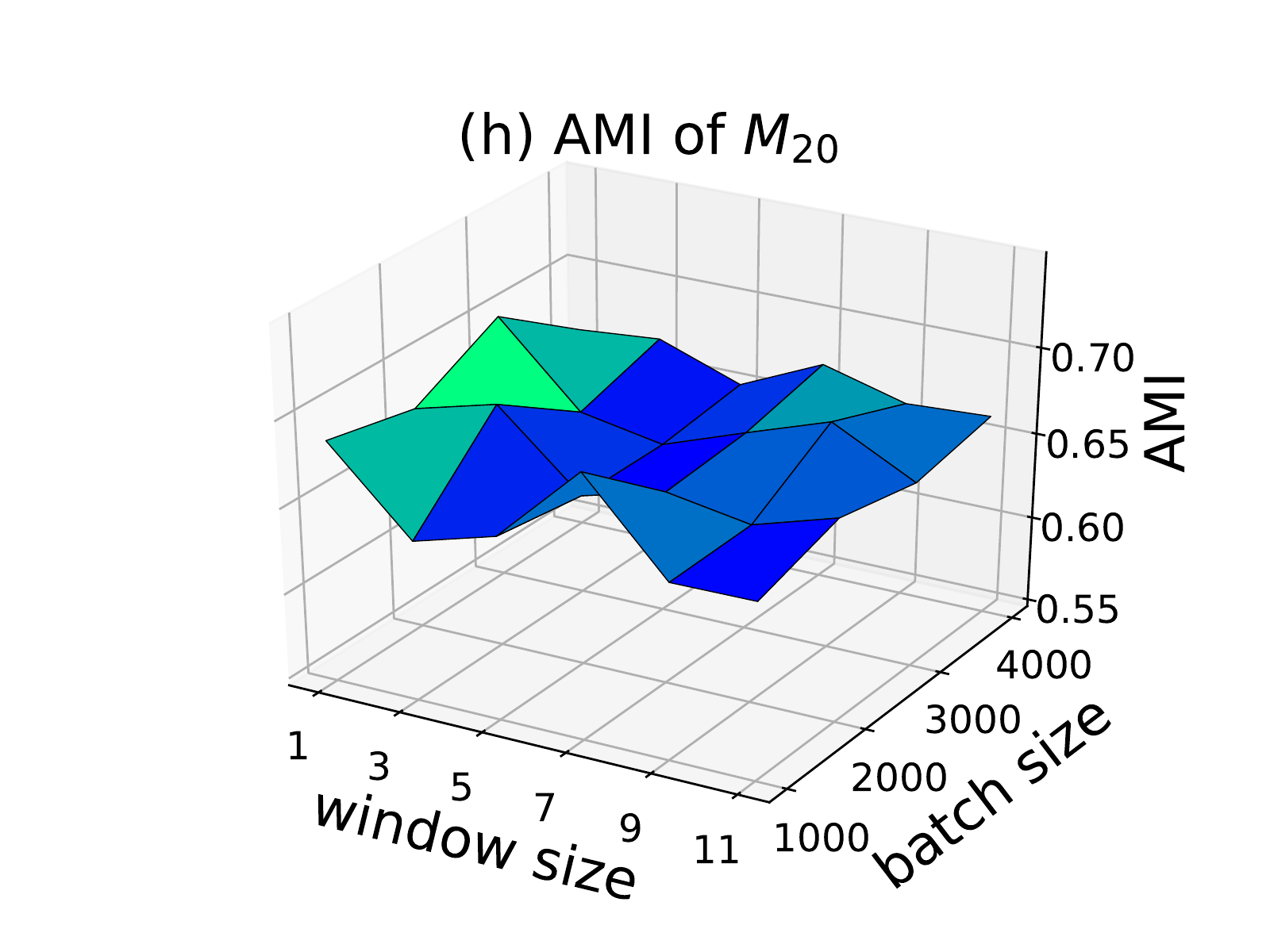}
    \end{minipage}
    }
    \centering
    {
    \begin{minipage}[c]{0.24\textwidth}
    \centering
        \includegraphics[width =4.3cm]{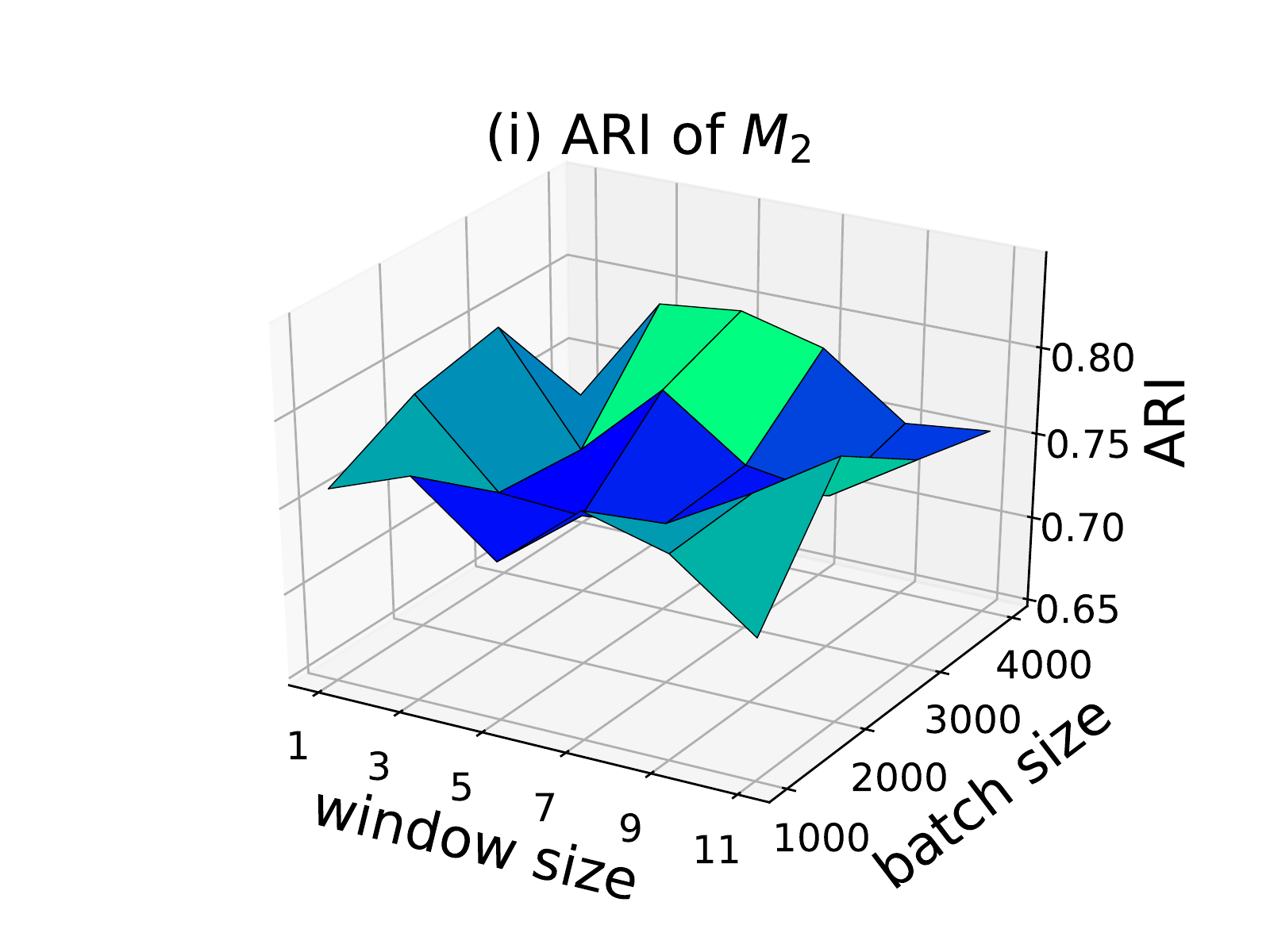}
    \end{minipage}
    }
    \centering
    {
    \begin{minipage}[c]{0.24\textwidth}
    \centering
        \includegraphics[width =4.3cm]{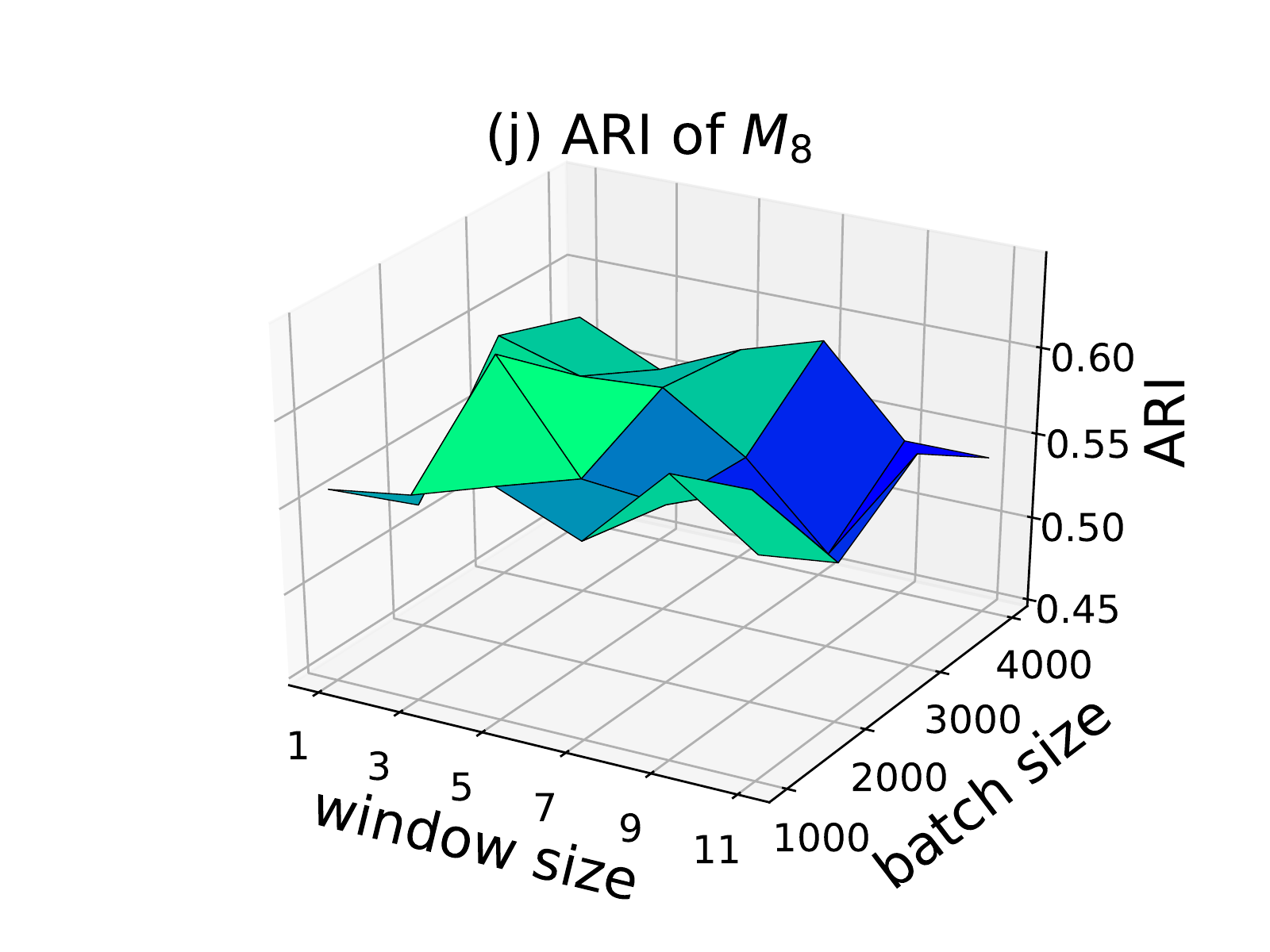}
    \end{minipage}
    }
    \centering
    {
    \begin{minipage}[c]{0.24\textwidth}
    \centering
        \includegraphics[width =4.3cm]{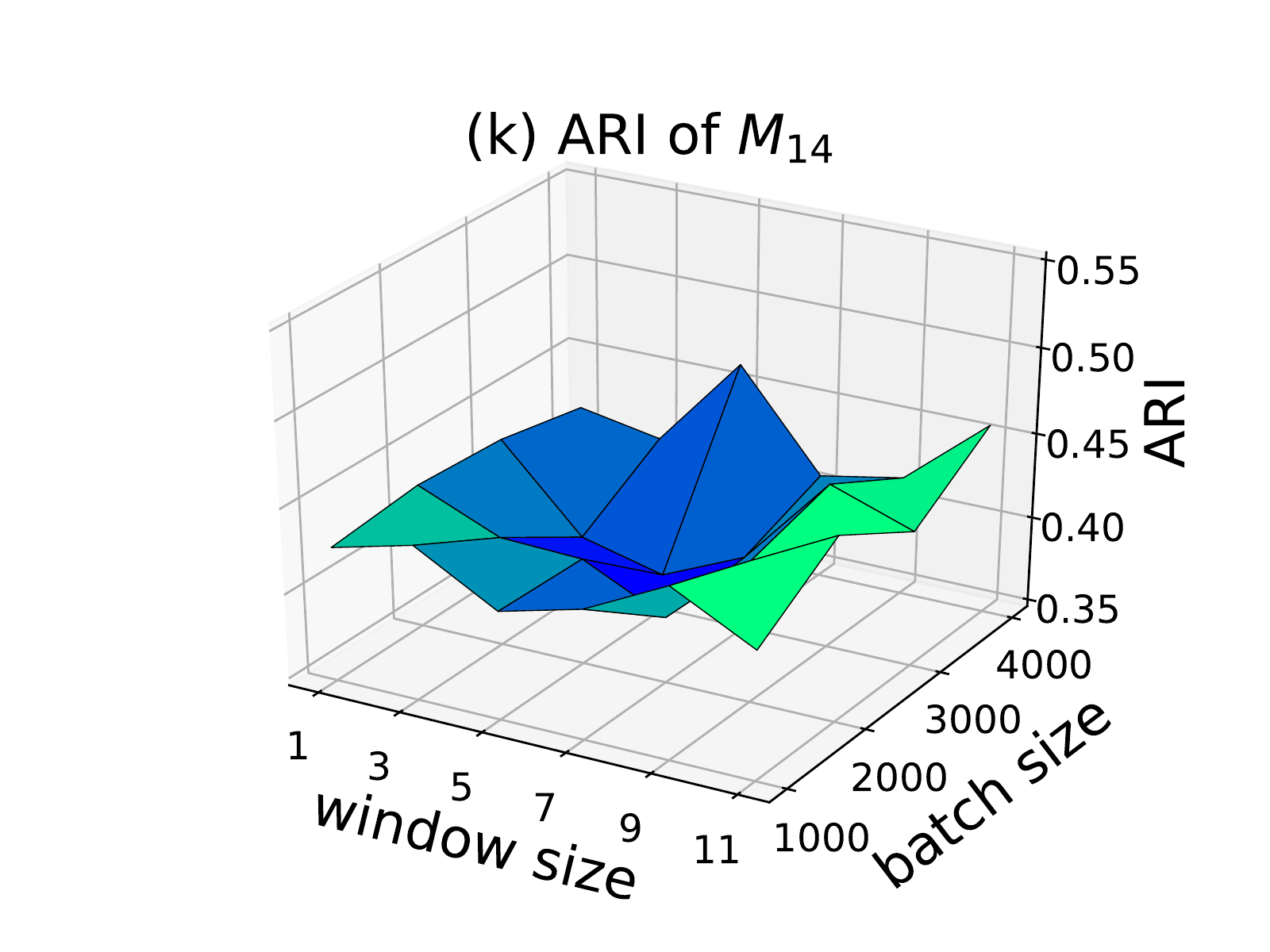}
    \end{minipage}
    }
    {
    \begin{minipage}[c]{0.24\textwidth}
    \centering
        \includegraphics[width =4.3cm]{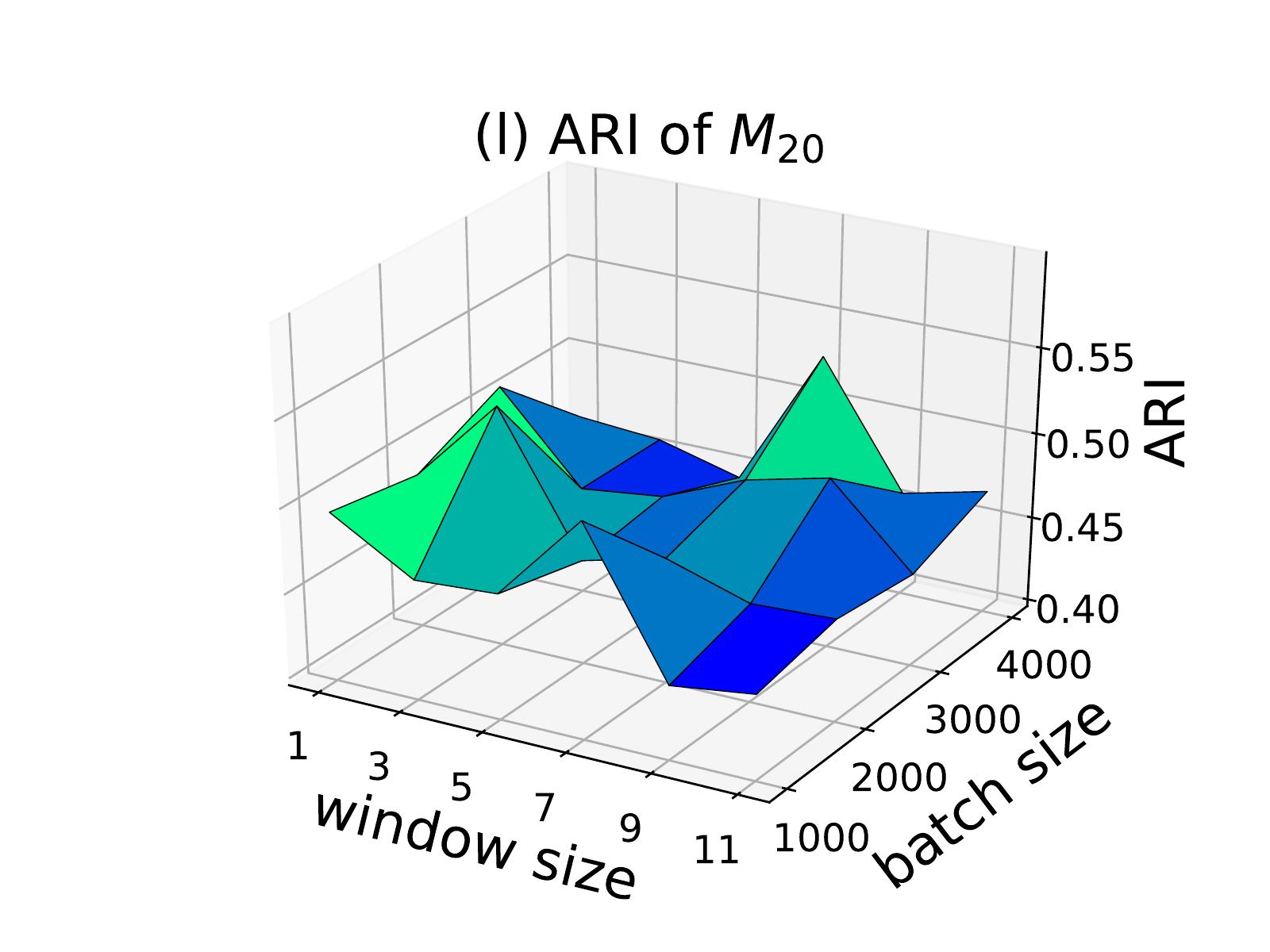}
    \end{minipage}
    }
    \caption{KPGNN with different hyperparameters. \textmd{We show the performance of KPGNN on message blocks $M_2$, $M_8$, $M_{14}$, and $M_{20}$ when adopting different window sizes and mini-batch sizes. \textbf{(a)}-\textbf{(d)} show the NMIs, \textbf{(e)}-\textbf{(h)} show the AMIs, and \textbf{(i)}-\textbf{(l)} show the ARIs. The colors indicate the fluctuations in values: the sunken areas are colored in blue and the convex areas in green.}}
    \label{fig:hyperparameter}
    \Description[KPGNN is insensitive to the changes in hyperparameters]{Twelve three dimensional sub-graphs. sub-graphs a to d, e to h, and i to l show the NMI, AMI, and ARI of KPGNN on message blocks 2, 8, 14, and 20. Each sub-graph has window sizes from 1 to 11 in increments of 2 on the X axis, batch sizes from 1,000 to 4,000 in increments of 1,000 on the Y axis, and the specific metric on the Z axis. The sub-graphs either show little fluctuations or some fluctuations but in a manner that is not clearly related to the changes in window size and batch size.}
\end{figure*}

\section{Experiments}\label{sec:experiment}
In this section, we first introduce the experiment setups, then compare KPGNN to baselines including offline as well as incremental social event detection models.
We also investigate the effects of adopting different forgetting strategies in the maintenance stage of KPGNN's life-cycle. 
Last, we provide a sensitivity analysis for the hyperparameters of KPGNN.

\subsection{Experiment Setup}
\subsubsection{Datasets}
We conduct experiments on two large-scale, publicly available datasets, i.e., the Twitter dataset \cite{mcminn2013building} and the MAVEN dataset \cite{wang2020MAVEN}. The Twitter dataset is collected to evaluate the DP social event detection methods. After filtering out repeated and irretrievable tweets, the dataset contains 68,841 manually labeled tweets related to 503 event classes, spread over a period of four weeks. MAVEN is a general domain event detection dataset constructed from Wikipedia documents. We remove sentences (i.e., messages) that are associated with multiple event types. The filtered dataset contains 10,242 messages related to 154 event classes.

\subsubsection{Baselines}
We compare KPGNN to general message representation learning and similarity measuring methods, offline social event detection methods, and the incremental ones. 
The baselines are: \textbf{Word2vec} \cite{mikolov2013efficient}, which uses the average of the pre-trained Word2vec embeddings of all the words in a message as its representation; 
\textbf{LDA} \cite{blei2003latent}, a generative statistical model that learns message representations by modeling the underlying topic and word distributions; 
\textbf{WMD} \cite{kusner2015word}, which measures the dissimilarity between two messages by calculating the minimum distance that the word embeddings in one need to travel to reach that of the other; \textbf{BERT} \cite{devlin2018bert}, which uses the average of BERT embeddings of all the words in a message as its representation;
\textbf{BiLSTM} \cite{graves2005framewise}, which learns bidirectional long-term dependencies between words in a message; \textbf{PP-GCN} \cite{peng2019fine}, an offline fine-grained social event detection method based on GCN\cite{kipf2016semi};
\textbf{EventX} \cite{liu2020story}, a fine-grained event detection method based on community detection and is applicable to the online scenario; 
\textbf{KPGNN$_t$}, a variation of the proposed KPGNN, in which the global-local pair loss term $\mathcal{L}_p$ is removed from the loss function and only the triplet loss term $\mathcal{L}_t$ is used.

\subsubsection{Experiment Setting}
For LDA, we set the total number of topics to 50. 
For EventX, we adopt the hyperparameters as suggested in the original paper \cite{liu2020story}. 
For BiLSTM and the GNN-based methods (PP-GCN, KPGNN, and KPGNN$_t$), we set the number of heads to $4$, embedding dimension $d'$ to $32$, the total number of layers $L$ to $2$, the learning rate to $0.001$, the optimizer to Adam, and the training epochs to $100$ with a patience of $5$ for early stopping. 
For KPGNN and KPGNN$_t$, we set the maintenance window $w$ to $3$, mini-batch size $|\{m_b\}|$ to $2000$, triplet margin $a$ to $3$, and the number of neighbors sampled for each message in the first layer $c_1$ and that of the second layer $c_2$ to $800$. 
We observe the effects of changing $w$ and $|\{m_b\}|$ in Section \ref{section:hyperparameter}. 
In incremental evaluation (Section \ref{section:online_eval}), we adopt the \textit{latest message strategy} (different update-maintenance strategies are detailed and studied in Section \ref{section:maintenance_eval}). 
We repeat all experiments $5$ times and report the mean and standard variance of the results. 
Note that although KPGNN does not require pre-defining the total number of event classes, some baselines (Word2vec, LDA, WMD, BERT, and BiLSTM) do. 
For a fair comparison, after obtaining the message similarity matrix from WMD and message representations from the other models except EventX (EventX does not pre-define its total number of detected classes), we leverage Spectral \cite{1238361} and K-Means clustering, respectively, and set the total number of classes to the number of ground-truth classes. Otherwise, DBSCAN \cite{ester1996density} can be used if the total number of classes is unknown as this is often the case of incremental detection.

For Word2vec, we use the pre-trained 300-d GloVe\cite{pennington2014glove} vectors \footnote{\url{https://spacy.io/models/en-starters##en_vectors_web_lg}}. 
For LDA, WMD, BERT, and PP-GCN, we use the open-source implementations \footnote{\url{https://radimrehurek.com/gensim/models/ldamodel.html}}\textsuperscript{,}\footnote{\url{https://tedboy.github.io/nlps/generated/gensim.similarities.html}}\textsuperscript{,}\footnote{\url{https://github.com/huggingface/transformers}}\textsuperscript{,}\footnote{\url{https://github.com/RingBDStack/PPGCN}}.
We implement EventX with Python $3.7.3$ and BiLSTM, KPGNN, and KPGNN$_t$ with Pytorch $1.6.0$. 
All experiments are conducted on a 64 core Intel Xeon CPU E5-2680 v4@2.40GHz with 512GB RAM and 1$\times$NVIDIA Tesla P100-PICE GPU.

\subsubsection{Evaluation Metrics}
To evaluate the performance of all models, we measure the similarities between their detected message clusters and the ground-truth clusters. We utilize normalized mutual information (NMI) \cite{estevez2009normalized}, adjusted mutual information (AMI) \cite{vinh2010information}, and adjusted rand index (ARI) \cite{vinh2010information}.
NMI measures the amount of information one can extract from the distribution of the predictions regarding the distribution of the ground-truth labels and is broadly adopted in social event detection method evaluations \cite{peng2019fine,liu2020story}.
AMI, similar to NMI, also measures the mutual information between two clusterings but is adjusted to account for chance \cite{vinh2010information}. 
ARI considers all prediction-label pairs and counts pairs that are assigned in the same or different clusters, and ARI also accounts for chance \cite{vinh2010information}.

\subsection{Offline Evaluation}
\label{section:offline_eval}
This subsection compares KPGNN to the baselines in an offline scenario. 
For both datasets, we randomly sample $70\%$, $20\%$, and $10\%$ for training, test, and validation, as such partition is commonly adopted by GNN studies \cite{peng2019fine}.

Tables \ref{table:offline_Twitter} and \ref{table:offline_MAVEN} summarize the results. 
KPGNN outperforms general message embedding methods (Word2vec, LDA, BERT, and BiLSTM) and similarity measuring methods (WMD) by large margins in all metrics (8$-$141$\%$, 4$-$1,200$\%$, and 29$-$2,100$\%$ in NMI, AMI, and ARI on the Twitter dataset and 13$-$49$\%$, 73$-$375$\%$, and 150$-$900$\%$ in NMI, AMI, and ARI on the MAVEN dataset). 
The reason for this is that these methods rely either on measuring the distributions of messages' elements (LDA) or message embeddings (Word2vec, WMD, BERT, and BiLSTM), and they all ignore the underlying social graph structure.
Different from them, KPGNN simultaneously leverages the semantics and structural information in the social messages and therefore acquires more knowledge.
KPGNN also outperforms both PP-GCN and KPGNN$_t$. 
This implies that introducing the global-local pair loss term $\mathcal{L}_p$ helps the model learn more knowledge from the graph structure.
Note that although PP-GCN shows strong performance, it assumes a stationary graph structure and cannot adapt to dynamic social streams. KPGNN, on the contrary, is capable of continuously adapting to and extending its knowledge from the incoming messages (empirically verified in Section \ref{section:online_eval}).
EventX shows higher NMI but much lower AMI and ARI compared to KPGNN. 
This suggests that EventX tends to generate a larger number of clusters, regardless of whether there is actually more information captured, while KPGNN is stronger in general, as it scores the highest or the second-highest in all metrics.

\subsection{Incremental Evaluation}
\label{section:online_eval}
This subsection evaluates KPGNN in an incremental detection scenario. 
We split the Twitter dataset by date to construct a social stream. Specifically, we use the messages of the first week to form an initial message block $M_0$ and the messages of the remaining days to form the following message blocks $M_1, M_2, ..., M_{21}$. 
Table \ref{table:social_stream} shows the statistics of the resulting social stream. 
Note that PP-GCN, as an offline baseline, cannot be directly applied to the dynamic social streams and we mitigate this by retraining a new PP-GCN model from scratch for each message block (i.e., use the previous blocks as the training set and predict on the current block).

Tables \ref{table:online_eval_nmi}, \ref{table:online_eval_ami} and \ref{table:online_eval_ARI} summarize the incremental social event detection results in NMI, AMI, and ARI, respectively. KPGNN significantly and consistently outperforms the baselines for all message blocks. 
KPGNN achieves relative performance gains over EventX by 6$-$27\% ($16\%$ on average) in NMI, 164$-$676\% ($319\%$ on average) in AMI, and 68$-$1782\% ($589\%$ on average) in ARI. 
The reason is, EventX relies solely on community detection, while KPGNN incorporates the semantics of the social messages. 
KPGNN outperforms WMD by 3$-$55\% ($22\%$ on average) in NMI, 3$-$48\% ($23\%$ on average) in AMI, and 3$-$68\% ($26\%$ on average) in ARI and over BERT by up to 19\% ($7\%$ on average) in NMI, up to 22\% ($8\%$ on average) in AMI, and up to 141\% ($58\%$ on average) in ARI. 
This is because KPGNN leverages the structural information of the social stream, which is ignored by WMD and BERT. 
KPGNN also outperforms PP-GCN by 38$-$67\% ($47\%$ on average) in NMI, 41$-$73\% ($53\%$ on average) in AMI, and up to 58\% ($27\%$ on average) in ARI. 
This suggests that KPGNN effectively preserves up-to-date knowledge, while PP-GCN can be distracted by obsolete information as the messages accumulate.
KPGNN generally outperforms KPGNN$_t$, which testifies the positive effect of incorporating more structural information by introducing the global-local pair loss term $\mathcal{L}_p$.
To conclude, the performance of KPGNN is superior to the baselines as it acquires and preserves more knowledge from the social messages.   

\subsection{Study on update-maintenance strategies}
\label{section:maintenance_eval}
Recall that KPGNN updates new messages into the message graph $\mathcal{G}$ in the detection stage (Figure \ref{fig:life_cycle} stage II). 
It also periodically removes obsolete messages from $\mathcal{G}$ and continues training to adapt to the new messages in the maintenance stage (Figure \ref{fig:life_cycle} stage III). 
The manner of updating and maintaining KPGNN affects its time complexity (discussed in Section \ref{section:complexity}), the knowledge it preserves, and, eventually, its performance. In this subsection, we compare \textit{three} different update-maintenance strategies, including: 

\textbf{1) All message strategy, keeping all the messages.} In the detection stage, we simply insert the newly arrived message block into $\mathcal{G}$. In the maintenance stage, we continue the training process using all the messages in $\mathcal{G}$. In other words, we let KPGNN memorize all the messages it ever received. This strategy is impractical (the messages accumulated in $\mathcal{G}$ can gradually slow down the model and will eventually exceed the embedding space capacity of the message encoder $\mathcal{E}$) and we implement it just for comparison purposes.
\textbf{2) Relevant message strategy, keeping messages that are related to the newly arrived ones.} In the detection stage, we insert the newly arrived message block into $\mathcal{G}$. In the maintenance stage, we first remove messages that are not connected to any messages that arrived during the last time window and then continue training using all the messages in $\mathcal{G}$. In other words, we let KPGNN forget the messages that are both old (i.e., arrived beyond the window) and unrelated (to the new messages that arrived within the window). Note that the knowledge acquired from these messages is preserved in the form of model parameters.
\textbf{3) Latest message strategy, keeping the latest message block.} In the detection stage, we use only the newly arrived message block to reconstruct $\mathcal{G}$. In the maintenance stage, we continue training with all the messages in $\mathcal{G}$, which only involves the latest message block. In other words, we let KPGNN forget all the messages except those in the latest message block. The knowledge learned from the removed messages is memorized in the form of model parameters.

Figures \ref{fig:strategy} (a)-(c) summarize the performance of KPGNN in NMI, AMI, and ARI when adopting the above three strategies in the incremental social event detection experiments. 
It is clear that the \textit{latest message strategy} achieves the strongest performance of all the strategies by discarding all messages in the past blocks while solely keeping the knowledge learned from these messages. 
Figures \ref{fig:strategy} (d) and (e) show the time and memory consumption of KPGNN when adopting these strategies. 
As expected, the \textit{latest message strategy} requires significantly less time and GPU memory compared to the others, as it keeps a light-weighted message graph $\mathcal{G}$. 
Note that the \textit{latest message strategy} and the \textit{relevant message strategy} consistently while the \textit{all message strategy} in most message blocks outperform strong baselines such as PP-GCN (the performance of PP-GCN is shown in Tables \ref{table:online_eval_nmi}-\ref{table:online_eval_ARI}). 
This proves the strong performance of KPGNN despite the update-maintenance strategies.

\subsection{Hyperparameter Sensitivity}
\label{section:hyperparameter}
This subsection studies the effects of changing $w$, the window size for maintaining KPGNN, and $|\{m_b\}|$, the mini-batch size in the incremental social event detection experiments. 
Figure \ref{fig:hyperparameter} compares the performance of KPGNN when adopting different window sizes and mini-batch sizes.
The NMI and AMI results in Figures \ref{fig:hyperparameter} (a)-(h) have small standard deviations in the range of 0.01-0.02.
This suggests that the NMIs and AMIs of KPGNN change with $w$ and $|\{m_b\}|$, but rather insignificantly. 
Adopting a smaller $w$ (1 or 3) in general gives a slightly better performance. 
For example, the block-wise average NMIs of window sizes 1 and 3 are 0.75 and 0.75, respectively, while that of window sizes 9 and 11 are 0.74 and 0.74, respectively. 
The mini-batch size also has little influence on the NMIs and AMIs. 
For example, the block-wise average NMIs of mini-batch sizes 1000, 2000, 3000, and 4000 are 0.75, 0.75, 0.75, and 0.75, respectively. 
In Figures \ref{fig:hyperparameter} (i)-(k), the ARIs of KPGNN show some fluctuations but in a manner that is not clearly related to the changes in the window size and the mini-batch size, as the block-wise average ARIs of the different window sizes range from 0.58-0.59 and that of the different mini-batch sizes range from 0.57-0.59. 
In a word, KPGNN is insensitive to the changes in hyperparameters.

\section{Related Work}
\label{section:related_work}
\textbf{Social Event Detection.} Based on their objectives, social event detection methods can be categorized into document-pivot (DP) methods \cite{liu2020story,aggarwal2012event,peng2019fine,hu2017adaptive,zhou2014event,zhang2007new} and feature-pivot (FP) ones \cite{fedoryszak2019real,fung2005parameter}. The former aim at clustering social messages based on their correlations while the latter aim at clustering social messages elements (such as words and named entities) based on their distributions. KPGNN is a DP method. Based on their application scenarios, social event detection methods can be divided into offline \cite{peng2019fine} and online \cite{fedoryszak2019real,liu2020story,hu2017adaptive,zhang2007new} ones. Though offline methods are essential in analyzing retrospective, domain-specific events such as disasters and political campaigns, online methods that continuously work on the dynamic social streams are desirable \cite{fedoryszak2019real}. Based on techniques and mechanisms, social event detection methods can be separated into popular classes such as methods rely on incremental clustering \cite{aggarwal2012event,ozdikis2017incremental,hu2017adaptive,zhang2007new}, community detection \cite{fedoryszak2019real,liu2020story,yu2017ring,liu2020event,liu2018event} and topic models \cite{zhou2014event}. These methods, however, learn limited amounts of knowledge as they ignore the rich semantics and structural information contained in the social streams to some extent. Furthermore, these models have too few parameters to preserve the learned knowledge. \cite{peng2019fine} proposes a GCN-based social event detection model, however, it can only work offline. KPGNN is different from the existing methods as it effectively acquires, extends, and preserves knowledge by continuously adapting to the incoming social messages.

As a side note, our work is different from \cite{deng2019learning} since 1) \cite{deng2019learning} addresses a different task, i.e., social event prediction, 2) \cite{deng2019learning} only uses the words for graph construction, while we utilize heterogeneous element types, and 3) \cite{deng2019learning} retrains a GCN from scratch at each time step while we continuously adapt to the incoming data.

\noindent
\textbf{Inductive Learning with Graph Neural Networks.} The past few years have witnessed the success of graph neural networks (GNNs) \cite{kipf2016semi,velivckovic2017graph,hamilton2017inductive} in graph data mining. In general, a GNN learns contextual node representations by extracting and aggregating local neighborhood information according to the input graph structure. Depending on their extraction and aggregation strategies, some GNNs \cite{kipf2016semi} only conduct transductive learning \cite{galke2019can} as they require pre-known, fixed graph structures. Others \cite{velivckovic2017graph,hamilton2017inductive} can be used in inductive learning \cite{galke2019can}, which means that they generalize to unseen nodes. Though often discussed, inductive learning using GNNs is rarely evaluated or utilized in real application scenarios \cite{galke2019can}. The proposed KPGNN is the first to leverage GNNs' inductive learning ability for incremental social event detection.

\section{Conclusion}
In this study, we address the task of incremental social event detection from a knowledge-preserving perspective. We design a novel KPGNN model that incorporates the rich semantics and structural information in social messages to acquire more knowledge. KPGNN continuously detects events and extends its knowledge using dynamic social streams. We empirically demonstrate the superiority of KPGNN compared to baselines through experiments.
A particularly interesting future research direction would be extending the proposed model for social event analysis (including studying the evolution of events) and causal discovery in social data.
\section*{Acknowledgement}
The authors of this paper were supported by NSFC through grants 62002007 and U20B2053, 
Key Research and Development Project of Hebei Province through grant 20310101D,
SKLSDE-2020ZX-12,
ARC DECRA Project through grant DE200100964,
NSF under grants III-1763325, III-1909323, and SaTC-1930941.

\bibliographystyle{ACM-Reference-Format}
\bibliography{reference}

\end{document}